\documentclass[10pt]{article}

\usepackage{etoolbox}
\newcommand{\arxiv}[1]{\iftoggle{iclr}{}{#1}}
\newcommand{\iclr}[1]{\iftoggle{iclr}{#1}{}}
\newtoggle{iclr}
\global\togglefalse{iclr}

\usepackage{microtype}
\usepackage{graphicx}
\usepackage{subcaption}
\usepackage{booktabs}
\usepackage{multirow}
\usepackage{multicol}
\usepackage{array}
\usepackage[dvipsnames]{xcolor}
\usepackage[most]{tcolorbox}
\usepackage{enumitem}
\usepackage{tabularx}
\usepackage{pifont}
\usepackage[symbol]{footmisc}

\setlist[itemize]{nosep,leftmargin=*,topsep=2pt}

\arxiv{
\usepackage[a4paper, margin=2.5cm]{geometry}
\PassOptionsToPackage{hypertexnames=false}{hyperref}  %
\usepackage[dvipsnames]{xcolor}
\usepackage[colorlinks]{hyperref}
\hypersetup{
    citecolor=[RGB]{50,100,170},
    linkcolor=[RGB]{50,100,170},
    urlcolor=[RGB]{255,102,178}}
\usepackage{fancyhdr}
}

\arxiv{

}

\makeatletter
\arxiv{
\renewcommand{\maketitle}{%
  \begin{center}%
    \vskip 0.35in%
    {\LARGE\bfseries \@title \par}%
    \vskip 0.3in%
    {\normalsize \@author \par}%
  \end{center}%
  \par
  \vskip 0.3in%
}

\renewcommand\section{\@startsection {section}{1}{\z@}{-2.0ex plus
    -0.5ex minus -.2ex}{1.5ex plus 0.3ex minus .2ex}{\large\bfseries\raggedright}}
\renewcommand\subsection{\@startsection{subsection}{2}{\z@}{-1.8ex plus
    -0.5ex minus -.2ex}{0.8ex plus .2ex}{\normalsize\bfseries\raggedright}}
\renewcommand\subsubsection{\@startsection{subsubsection}{3}{\z@}{-1.5ex plus
   -0.5ex minus -.2ex}{0.5ex plus .2ex}{\normalsize\bfseries\raggedright}}

\renewenvironment{abstract}%
  {\centerline{\large\bfseries Abstract}%
   \begin{list}{}%
      {\setlength{\rightmargin}{0.6cm}%
       \setlength{\leftmargin}{0.6cm}}%
    \item[]\ignorespaces}%
  {\unskip\end{list}}

\setlength{\parindent}{0pt}
\setlength{\parskip}{0.4em}
}
\makeatother

\iclr{
\usepackage[colorlinks=true, linkcolor=blue!70!black, citecolor=blue!70!black,urlcolor=blue!70!black,breaklinks=true]{hyperref}
\PassOptionsToPackage{dvipsnames}{xcolor} 
}

\usepackage{microtype}
\usepackage{hhline}

\makeatletter
\newcommand{\neutralize}[1]{\expandafter\let\csname c@#1\endcsname\count@}
\makeatother

\usepackage{algorithm}

\arxiv{
\usepackage{natbib}
\bibliographystyle{plainnat}
\bibpunct{(}{)}{;}{a}{,}{,}
}

\usepackage{amsthm}
\usepackage{mathtools}
\usepackage{amsmath}
\usepackage{bbm}
\usepackage{amsfonts}
\usepackage{amssymb}

\usepackage{xpatch}

\usepackage{thmtools}
\usepackage{thm-restate}
\declaretheorem[name=Theorem]{theorem}
\declaretheorem[name=Lemma,sibling=theorem]{lemma}
\declaretheorem[name=Assumption,sibling=theorem]{assumption}
\declaretheorem[name=Condition,sibling=theorem]{condition}

\declaretheorem[name=Proposition,sibling=theorem]{proposition}

\makeatletter
  \renewenvironment{proof}[1][Proof]%
  {%
   \par\noindent{\bfseries\upshape {#1.}\ }%
  }%
  {\qed\newline}
  \makeatother

\theoremstyle{definition}  %

\newtheorem{corollary}{Corollary}[section]

\theoremstyle{plain}
\newtheorem{definition}{Definition}[section]

\xpatchcmd{\proof}{\itshape}{\normalfont\proofnameformat}{}{}
\newcommand{\proofnameformat}{\bfseries}

\usepackage[nameinlink,capitalize]{cleveref}

\renewcommand{\eqref}[1]{\texorpdfstring{\hyperref[#1]{(\ref*{#1})}}{(\ref*{#1})}}

\crefformat{equation}{#2Eq. (#1)#3}
\Crefformat{equation}{#2Eq. (#1)#3}

\Crefformat{figure}{#2Figure #1#3}
\Crefname{assumption}{Assumption}{Assumptions}
\Crefformat{assumption}{#2Assumption #1#3}
\Crefname{subsubsection}{Section}{Sections}
\crefformat{subsubsection}{#2Section #1#3}
\Crefformat{subsubsection}{#2Section #1#3}
\Crefname{alg}{Alg.}{Algs.}

\usepackage{crossreftools}
\usepackage{xparse}

\ExplSyntaxOn
\DeclareDocumentCommand{\XDeclarePairedDelimiter}{mm}
 {
  \__egreg_delimiter_clear_keys: %
  \keys_set:nn { egreg/delimiters } { #2 }
  \use:x %
   {
    \exp_not:n {\NewDocumentCommand{#1}{sO{}m} }
     {
      \exp_not:n { \IfBooleanTF{##1} }
       {
        \exp_not:N \egreg_paired_delimiter_expand:nnnn
         { \exp_not:V \l_egreg_delimiter_left_tl }
         { \exp_not:V \l_egreg_delimiter_right_tl }
         { \exp_not:n { ##3 } }
         { \exp_not:V \l_egreg_delimiter_subscript_tl }
       }
       {
        \exp_not:N \egreg_paired_delimiter_fixed:nnnnn 
         { \exp_not:n { ##2 } }
         { \exp_not:V \l_egreg_delimiter_left_tl }
         { \exp_not:V \l_egreg_delimiter_right_tl }
         { \exp_not:n { ##3 } }
         { \exp_not:V \l_egreg_delimiter_subscript_tl }
       }
     }
   }
 }

\keys_define:nn { egreg/delimiters }
 {
  left      .tl_set:N = \l_egreg_delimiter_left_tl,
  right     .tl_set:N = \l_egreg_delimiter_right_tl,
  subscript .tl_set:N = \l_egreg_delimiter_subscript_tl,
 }

\cs_new_protected:Npn \__egreg_delimiter_clear_keys:
 {
  \keys_set:nn { egreg/delimiters } { left=.,right=.,subscript={} }
 }

\cs_new_protected:Npn \egreg_paired_delimiter_expand:nnnn #1 #2 #3 #4
 {%
  \mathopen{}
  \mathclose\c_group_begin_token
   \left#1
   #3
   \group_insert_after:N \c_group_end_token
   \right#2
   \tl_if_empty:nF {#4} { \c_math_subscript_token {#4} }
 }
\cs_new_protected:Npn \egreg_paired_delimiter_fixed:nnnnn #1 #2 #3 #4 #5
 {
  \mathopen{#1#2}#4\mathclose{#1#3}
  \tl_if_empty:nF {#5} { \c_math_subscript_token {#5} }
 }
\ExplSyntaxOff

\XDeclarePairedDelimiter{\supnorm}{
  left=\lVert,
  right=\rVert,
  subscript=\infty
  }
 
\newcommand{\CompanyName}{Datadog}
\let\mathuparrow\uparrow
\renewcommand{\uparrow}{\ensuremath{\mathuparrow}}
\newcommand{\cmark}{\textcolor{OliveGreen}{\ding{51}}}
\newcommand{\xmark}{\textcolor{red}{\ding{55}}}

\newtcolorbox{outercasebox}[2][]{%
  enhanced,
  colback=white,
  colframe=black!25,
  boxrule=0.6pt,
  arc=2pt,
  left=8pt,right=8pt,top=8pt,bottom=8pt,
  boxsep=1pt,
  before skip=4pt, after skip=4pt,
  fonttitle=\bfseries,
  title={#2},
  minipage,
  valign=top,
  #1
}
\newtcolorbox{casebox}[2][]{%
  enhanced,
  breakable,
  colback=white,
  colframe=black!25,
  boxrule=0.6pt,
  arc=2pt,
  left=6pt,right=6pt,top=4pt,bottom=4pt,
  boxsep=1pt,
  before skip=4pt, after skip=4pt,
  fonttitle=\bfseries,
  title={#2},
  #1
}
\newtcolorbox{answerbox}[2][]{%
  enhanced,
  colback=#2!7,
  colframe=#2!55,
  boxrule=0.8pt,
  arc=2pt,
  left=6pt,right=6pt,top=4pt,bottom=4pt,
  boxsep=1pt,
  before skip=0pt, after skip=0pt,
  #1
}

\title{\textsc{ARFBench}: Benchmarking Time Series Question Answering Ability for Software Incident Response}

\author{
Stephan Xie$^{1,2}$, Ben Cohen$^2$, Mononito Goswami$^3$\footnote[1]{This work does not relate to the author's position at Amazon.}, Junhong Shen$^1$,\\
Emaad Khwaja$^2$, Chenghao Liu$^2$, David Asker$^2$, Othmane Abou-Amal$^2$, Ameet Talwalkar$^{1,2}$\\[0.5em]
$^1$Machine Learning Department, Carnegie Mellon University, Pittsburgh, PA, USA\\
$^2$Datadog AI Research, New York, NY, USA\\
$^3$Amazon Web Services, Seattle, WA, USA\\[0.25em]
\texttt{\{stephan.xie, ameet.talwalkar\}@datadoghq.com}
}

\begin{document}

\maketitle

\begin{abstract}
Time series question-answering (TSQA), in which we ask natural language questions to infer and reason about properties of time series, is a promising yet underexplored capability of foundation models.
In this work, we present \textsc{ARFBench}, a TSQA benchmark that evaluates the understanding of multimodal foundation models (FMs) on time series anomalies prevalent in software incident data. \textsc{ARFBench} consists of 750 questions across 142 time series and 5.38M data points from 63 production incidents sourced exclusively from internal telemetry at \CompanyName{}. We evaluate leading proprietary and open-source LLMs, VLMs, and time series FMs and observe that frontier VLMs perform markedly better than existing baselines; the leading model (GPT-5) achieves a 62.7\% accuracy and 51.9\% F1.
We next demonstrate the promise of specialized multimodal approaches. We develop a novel TSFM + VLM hybrid prototype which we post-train on a small set of synthetic and real data that yields comparable overall F1 and accuracy with frontier models. Lastly, we find models and human domain experts exhibit complementary strengths. We define a model-expert oracle, a best-of-2 oracle selector over model and expert answers, yielding 82.8\% F1 and 87.2\% accuracy and establishing a new superhuman frontier for future TSQA models. 
The benchmark is available at \href{https://huggingface.co/datasets/Datadog/ARFBench}{https://huggingface.co/datasets/Datadog/ARFBench}. \end{abstract}

\section{Introduction}
Trillions of dollars are lost each year due to system outages \citep{krasner2018cost}. 
In order to resolve unexpected software incidents in a timely manner, engineers must decide on their next actions and determine where errors occur.
To guide these choices, engineers rely on \textit{observability metrics}, or 
time series used to monitor the health of production software, 
such as CPU usage, latency, and error counts, among others. Engineers investigate incidents by repeatedly asking questions about these time series, such as \textit{``Is the pattern in this time series abnormal?"}, or \textit{``Are these two time series correlated?"}.
These questions naturally give rise to time series question answering (TSQA) tasks, in which reasoning over multiple time series and temporal patterns is necessary in order to support downstream decision-making. Despite the central role of such question-driven analysis in incident response, it remains unclear whether modern foundation models can reliably answer the kinds of time series questions engineers ask in practice.

\begin{figure*}[t]
  \centering
    \includegraphics[width=\linewidth]{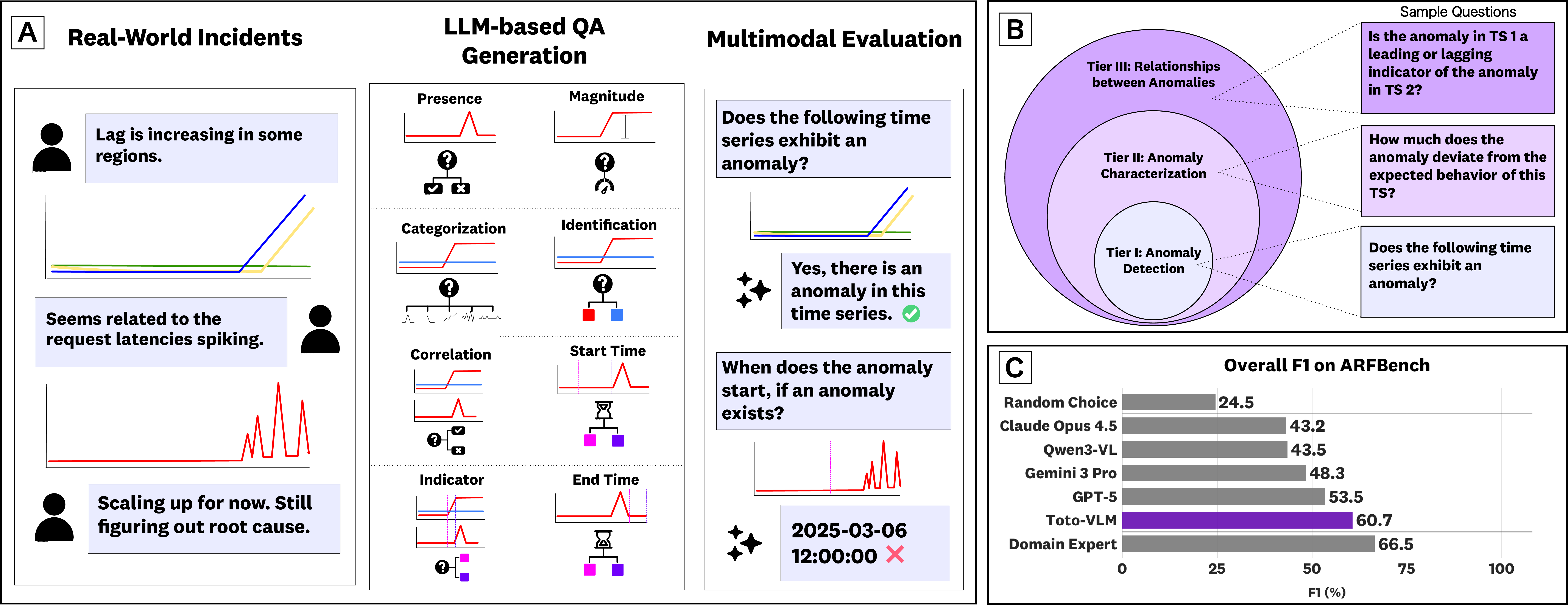}
        \caption{\textsc{ARFBench} consists of 750 question-answer (QA) pairs, derived from 63 real-world incidents and 142 observability time series. 
Observability time series are highly nonstationary and complex \citep{cohen2025time}, and \textsc{ARFBench} includes highly multivariate series that challenge LLM/VLM input representations. \textbf{A.} 
Workflow of ARFBench question-answer generation. Engineers use commercial messaging platforms to respond to incidents, where they typically send time series widgets that visualize relevant metrics. Time series and incident timelines from internally monitored incidents are used as input to an LLM pipeline and fit to eight different question templates testing various aspects of anomalies. The resulting multiple choice question-answer pairs can be used to evaluate various predictive models. \textbf{B.} ARFBench questions are designed to capture escalating levels of difficulty. \textbf{C.} Our model, Toto-1.0-QA-Experimental, obtains high accuracy and F1 on ARFBench, comparable to recent frontier models.}
  \label{fig:summary-figure}
\end{figure*}

To this end, we present the \textsc{A}nomaly \textsc{R}easoning \textsc{F}ramework \textsc{Bench}mark (\textsc{ARFBench}), a multimodal, multiple-choice question-answering benchmark designed to test time series reasoning and understanding tasks that are constructed from real software incidents sourced exclusively from internal telemetry at \CompanyName{}, a leading observability platform vendor, with sensitive data anonymized by both an LLM and manual review. ARFBench is distinguished by three key aspects: (i) it is grounded in real production time series from operational systems, (ii) each QA pair is supported by expert annotations and additional incident context, and (iii) tasks are designed to evaluate compositional anomaly reasoning, with questions organized into three tiers of increasing difficulty, with higher-tier tasks depending on correct reasoning over lower tiers. 

ARFBench questions are designed in tiers that capture increasing levels of reasoning: first detecting anomalies, then identifying properties of anomalies, and lastly reasoning about the relationships between anomalies across time series. These questions test general time series understanding abilities such as event classification, magnitude estimation, trend and periodicity analysis, and also require robust understanding of correlation and temporal relationships (Section \ref{sec: problem-definition}). \looseness=-1

We benchmark a wide variety of foundation models on ARFBench, including leading proprietary and open-source LLMs, vision-language models (VLMs), and time series LLMs (Section \ref{sec:baselines}). However, we find that the high number of variates in ARFBench data poses major problems for plotting and context-length limits in both LLMs and VLMs, motivating a need for specialized time series modeling.
Therefore, we also devise novel hybrid modeling methods by combining time series foundation models (TSFMs) with VLMs. 

We observe three key takeaways from our experiments.
\begin{enumerate}[wide, labelwidth=!, labelindent=0pt]
    \item \textbf{Frontier VLMs Markedly Lead Over Existing Baselines.} GPT-5, the top pre-existing model we evaluate, achieves 51.9\% macro F1 and 62.7\% accuracy, surpassing naive baselines such as the random choice baseline or the per-category majority predictor by at least 17 percentage points in accuracy and F1. Moreover, our results corroborate past works on time series classification \citep{daswani2024plots}, demonstrating that VLMs tend to outperform LLMs.
    \item \textbf{Hybrid TSFM-VLM Models Are a Promising Direction For Specialized TSQA Modeling.} While hybrid TSFM-LLM approaches have previously been explored, they have only been compared against LLMs or weaker VLMs as a baseline. Our post-trained TSFM-VLM model achieves comparable overall performance in terms of accuracy and F1 with frontier models. We hypothesize that the joint training of time series encoders and VLMs is crucial for improving model performance on time series QA.
    \item \textbf{Models and Domain Experts Demonstrate Complementary Abilities, Establishing a New Superhuman Frontier.} We observe substantially different error profiles between leading models and human experts, suggesting that their strengths are complementary. We introduce a model-expert oracle which yields 82.8\% F1 and 87.2\% accuracy, establishing a new superhuman frontier for future TSQA models.
\end{enumerate}
Additionally, we analyze the impact of multimodal inputs and training stages on model performance and provide qualitative error analyses across models (Section \ref{app: ablations} and Appendix \ref{app:case-studies}). Our results suggest a promising avenue for future TSQA models, especially for observability and software incident response.

We release the benchmark data \href{https://huggingface.co/datasets/Datadog/ARFBench}{here}. We also maintain a public leaderboard \href{https://huggingface.co/spaces/Datadog/ARFBench}{here}. Furthermore, the code is located \href{https://github.com/datadog/arfbench}{here}.

\begin{table*}[t]
\centering
\resizebox{\textwidth}{!}{%
\begin{tabular}{lcccccc}
\toprule
\textbf{Benchmark} & \textbf{Domain} &\textbf{Real-World}  &\textbf{Context-Enriched}& \textbf{Multivariate}   &\textbf{Expert-Annotated}& \textbf{Reasoning over Multiple Time Series}\\
\midrule
 ChatTime TSQA \cite{wang2025chattime} & Agnostic& \xmark& \xmark& \xmark &\xmark& \xmark\\
TimeSeriesExam \cite{cai2024timeseriesexam}& Agnostic & \xmark&\cmark& \xmark   &\xmark& \cmark\\
ECG-QA \cite{oh2023ecg} & Medical & \cmark  &\xmark& \cmark&\cmark& \cmark  \\
 ChatTS \cite{xie2024chatts} & AIOps & \xmark& \cmark& \cmark &\xmark& \cmark\\
 Time-MQA \cite{kong2025time} & Multi-domain & \cmark  & \cmark & \xmark  &\xmark& \xmark\\
 TSR-Suite \cite{guan2025timeomni} & Multi-domain& \cmark& \cmark& \xmark &\xmark&\cmark\\
 EngineMT-QA \cite{wang2025itformer} & Aerospace& \xmark& \xmark& \cmark &\cmark&  \cmark\\
 Time-RA \cite{yang2025time}& Multi-domain& \cmark& \cmark& \cmark &\xmark&\xmark\\
 Telecom-TS \cite{feng2025telecomts}& Observability& \cmark& \cmark& \cmark &\xmark&\xmark\\
\textsc{ARFBench}     & Observability  & \cmark  &\cmark& \cmark   & \cmark& \cmark  \\
\bottomrule
\end{tabular}
}
\vspace{1mm}
\caption{Comparison between existing time series QA benchmarks. \textsc{ARFBench} is a benchmark with real world, multivariate, context-enriched time series, and is grounded in expert annotations. Moreover, \textsc{ARFBench} also evaluates reasoning over multiple time series, which is an important task for software incident response. \textbf{Columns:} \textit{Real-world} refers to collected or measured time series data in the application domain, as opposed to synthetic or simulated data. \textit{Context-Enriched} refers to text input additional to (but not including) the question. \textit{Multivariate }refers to time series with more than one channel. \textit{Expert-Annotated} refers to human expert-authored primary source evidence analyzing the time series data. \textit{Reasoning over Multiple Time Series} refers to question-answer pairs that include two or more different time series within the question. 
}
\label{tab:qa-benchmark-checks}
\end{table*} \section{Related Work}\label{sec: existing-work}
\textbf{Evaluating Multimodal Time Series Reasoning and QA.} 
Existing works explore multimodal time series understanding through classical tasks such as forecasting \cite{williams2024context, liu2024time, nasbench360} or in chart analysis \cite{masry2022chartqa, xu2023chartbench, shen2025cat}. 
Recently, TSQA benchmarks have emerged as a test of reasoning ability. Synthetic TSQA benchmarks \cite{merrill2024language, cai2024timeseriesexam, anonymous2026timeseriesexamagent, wang2025itformer, divo2025quants, xie2024chatts}, where the time series is synthetic, have been a popular benchmarking approach.
Real-world TSQA benchmarks are less common and often are derived from pre-existing benchmarks or datasets: ECG-QA \cite{oh2023ecg}, TS-Insights \cite{zhang2023insight}, Time-mqa \cite{kong2025time}, TimerBed \cite{liu2025picture}, and TSR-Suite \cite{guan2025timeomni} generate QA pairs over diverse domains using publicly available time series data, but lack reasoning complexity in multivariate or multi-time series analysis and often use synthetic anomaly injection. Furthermore, many real-world TSQA datasets derive synthetic question-answer pairs without contextual evidence, i.e. there are no auxiliary human annotations that act as primary evidence for the proposed labels. Among existing benchmarks, TelecomTS \cite{feng2025telecomts} and Time-RA \cite{yang2025time} are the most similar to this work, but neither benchmark evaluates on reasoning over multiple time series, nor are they grounded in expert annotations. Moreover, existing benchmarks often evaluate in a single-task framework, training separate models for each task. 
In contrast, \textsc{ARFBench} tests TSQA ability over eight different anomaly-related reasoning tasks with observability time series derived from real software incidents. Enriched with context and human annotations, the time series and QA pairs in ARFBench present a novel challenge in TSQA and reasoning, grounded in real-world and complex data.

\textbf{Time Series Anomaly Detection.} Time series anomaly detection is the task of proactively predicting or reactively finding significant deviations from expected behavior and trends in time series \cite{chandola2009anomaly, li2025codepde, xu2024specialized, goswami2022unsupervised}. Many benchmarks, including those grounded in observability, have been developed to assess anomaly detection \cite{lavin2015evaluating, ahmad2017unsupervised, han2022adbench, schmidl2022anomaly, si2024timeseriesbench, yang2025time}. Our work is related to, but fundamentally distinct from, classical anomaly detection benchmarks.
Traditional anomaly detection formulations typically require models to output precise anomalous time ranges, and their evaluation relies on specialized metrics that attempt to reconcile ambiguity in anomaly boundaries and downstream utility \citep{huet2022local, garg2021evaluation, paparrizos2022volume, liu2024elephant}.
These benchmarks also commonly exhibit disagreement in ground-truth labeling, reflecting the inherent subjectivity in defining exact anomaly start and end times \citep{wu2021current}.
In contrast, ARFBench does not evaluate anomaly detection as a time-range prediction problem.
Instead, we focus on \emph{high-level anomaly reasoning} by framing incident analysis as a series of multiple-choice questions.
Specifically, we split the task into three separate single-class classification tasks (presence, start time, end time). In doing so, we penalize performance for incorrect answers equally and thus allow a set of time ranges to equally represent the valid time range of the anomaly; the model need only find the answer that is closest in time to the ground truth. This simplified task measures reasoning relevant to downstream decisions in incident response, where exactness of time ranges is usually not necessary. Beyond classical anomaly detection, ARFBench also includes tasks that require reasoning about anomaly types and relationships between multiple time series, such as anomaly categorization and cross-metric anomaly correlation. \looseness=-1

\textbf{Modeling Methods in TSQA.} 
Recent works have studied how zero-shot VLMs and reinforcement learning (RL) post-trained VLMs \cite{chow2024towards,mixtureofmamba,daswani2024plots, sen2025bedtime, liu2025picture, zhang2025timemaster, he2025harnessing} perform on time series tasks such as time series classification, TSQA, and captioning on existing time series data. VLMs have also been used for understanding anomalies in synthetic time series data \cite{xie2024chatts, zhou2024can, shen2022dash, shen2024ups} as well as representation and reasoning in TSFMs \cite{potosnak2025investigating, wilinski2024exploring, yu2025ts}.
In parallel, multimodal time series LLM and VLM models have received increasing attention for forecasting \cite{zhong2025time, zhang2025tempogpt, kim_multi-modal_2024, cheng_financial_2022}, anomaly detection and classification \cite{zhang_dualtime_2024, ebrahimi_lanistr_2024, zhang2025timesense, lan2025axis}, and summarization \cite{cai_jolt_2024, zhang2025timesense}.
Unlike existing methods that train time series encoders from scratch, we utilize a novel approach to combine and train pretrained TSFMs with VLMs jointly, demonstrating promise in multimodal foundation model approaches to TSQA modeling. \textsc{ARFBench} is designed to allow evaluation of reasoning ability using vision, language, and time series data modalities within a well-structured TSQA framework.

\section{ARFBench}\label{sec:ARFBench}
In this section, we describe the problem and benchmark design, data sources, and data generation pipeline for \textsc{ARFBench}.
\subsection{Problem Definition}
\label{sec: problem-definition}
\begin{figure*}
    \centering
    \includegraphics[width=\linewidth]{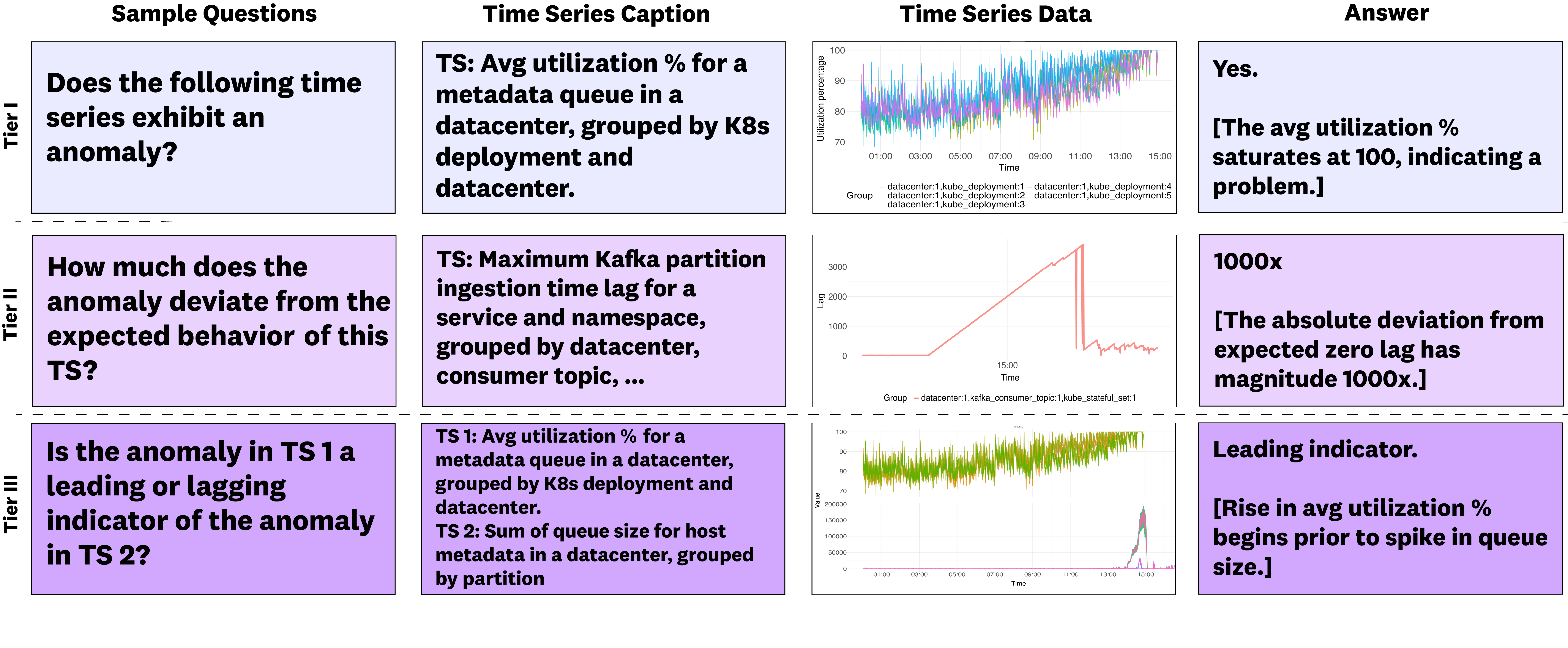}
    \caption{Example questions in ARFBench for each tier. For each question, the model is given a time series caption along with corresponding time series data in image or tabular form. In addition to the input above, time series channels are named with tags that represent collections such as datacenters, pods, services, and more.}
    \label{fig:example-questions}
\end{figure*}
Each question in \textsc{ARFBench} consists of a templated question prompt, a caption describing one or two univariate or multivariate time series of interest, and the associated time series data. In each time series, there may be zero, one, or multiple anomalies.

The 8 question categories are designed to test escalating levels of understanding (Figure \ref{fig:summary-figure}). By design, as the difficulty of the question increases, the questions require more complex reasoning and additional multimodal evidence. We assign each category to a tier under this framework. The Tier I difficulty (\textit{Presence}) consists of binary decisions that identify the presence or absence of an anomaly.
The Tier II difficulty (\textit{Identification, Magnitude, Start Time, End Time, Categorization}) corresponds to questions that require analysis of a single (possibly multivariate) time series in order to characterize specific properties of an anomaly or identify anomalous channels, thus necessitating reasoning about the time range and series values.
Tier III difficulty (\textit{Correlation, Leading/Lagging Indicator}) targets paired time series reasoning. In addition to Tier I and II-style detection and localization within each series, the model must compare anomaly structure across series—including temporal alignment and relative ordering—to determine whether anomalies are related and, if so, whether one series tends to lead or lag the other. More details about question design can be found in Appendix \ref{app: question-defs}.

\begin{figure}
    \centering
    \includegraphics[width=\linewidth]{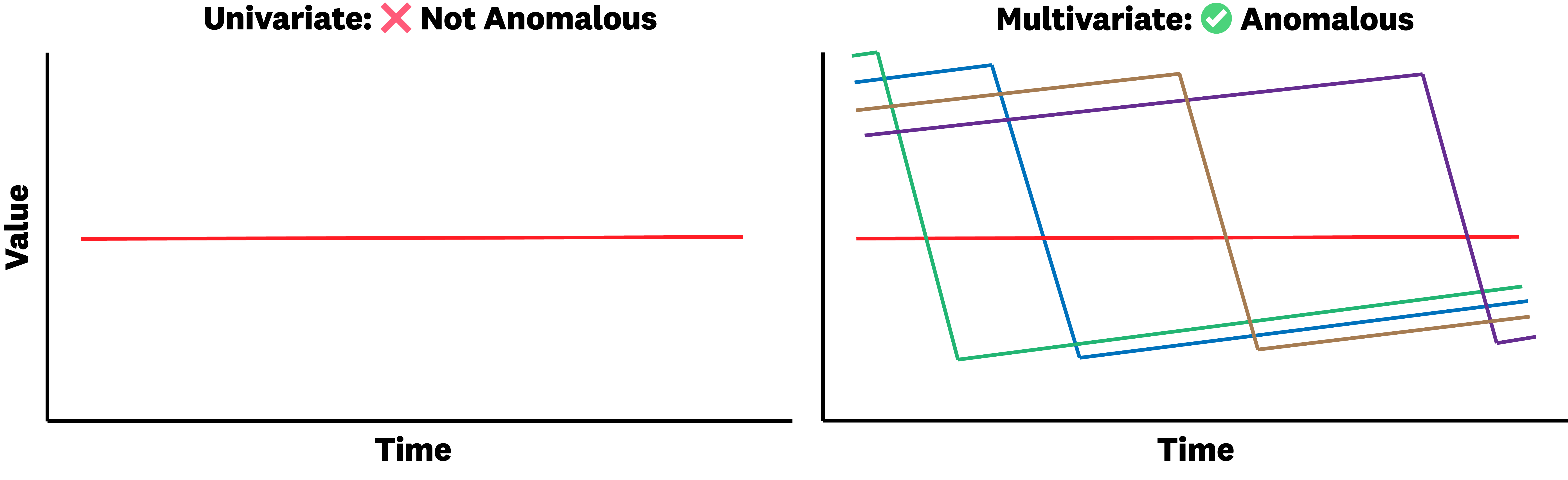}
    \caption{ARFBench requires multivariate reasoning. A single time series variate taken out of context may not appear to be anomalous. However, in a structured multivariate grouping, the variate would be considered anomalous. The time series shown above is based on average TLS certificate lifetime across cluster IDs.}
    \label{fig:multivariateness}
\end{figure}
\subsection{Key Design Decisions}
\textbf{Defining Anomaly Understanding as a Classification Problem. } 
The questions in \textsc{ARFBench} all focus on understanding time series anomalies, and therefore it is important for us to define anomalies in a consistent manner.

Historically, anomalies have been defined around scoring either points or subsequences of the time series with anomaly scores \cite{chandola2009anomaly}. Evaluation of anomaly detection has proven to be difficult, both from the benchmark and methodological side, since ambiguities exist both in terms of the ground truth (e.g. where should the start of an anomaly be?) and the evaluation metrics (e.g. how should one penalize the start of a predicted anomaly subsequence based on its distance from the start of the ground truth?). Improving evaluations in anomaly detection is still an active area of research (see Section \ref{sec: existing-work} for more details). Here, we choose to sidestep issues with ground truth determination and choice of measure by using a multiple-choice format. In doing so, we eliminate boundary problems by constructing answer choices with a single correct choice relative to the rest.

\textbf{Incorporating Additional Context is Necessary For Correct Reasoning.} 
Anomalies are inherently contextual: whether a pattern is considered anomalous depends not only on the observed values, but also on temporal context, the semantic meaning of the signal, and its relationships to other time series. As additional context becomes available—such as longer historical windows or richer metadata about what a signal represents—the interpretation of anomalous behavior can fundamentally change. A small spike in the time series that measures unauthorized access attempts for a service may be much more severe than a brief spike in the latency of the same service. The former could suggest compromised systems, while the latter may be transient noise, with little long-term impact. Therefore, an important component of detecting and analyzing anomalies is understanding what the time series represents. In \textsc{ARFBench}, we include this crucial context by summarizing the queries used to fetch individual time series examples from a database (Figure \ref{fig:query_summary}, Appendix).

Moreover, multivariate groupings (Figure \ref{fig:multivariateness}) provide insight into whether a time series variate is anomalous or not. Although multivariate anomaly detection could be done by analyzing each variate individually, this misses the broader relationship of the variate to the multivariate grouping, if structure exists. For example, a service with one pod failing and restarting may not be significant, but a combination of pods in the same service failing and restarting simultaneously may be a critical error. In \textsc{ARFBench}, all time series variates are grouped under meaningful structure, e.g. variates are grouped across specific services, datacenters, or other attributes. This requires TSQA models to reason about anomalies in a multivariate rather than univariate fashion.

\subsection{Software Incident Timelines as Expert Annotations}
\label{sec:incident-timelines}
\textbf{Software Incidents.} Formally, software incidents are unplanned degradations or failures of a software system that deviate from its specified or expected behavior. Diagnosing such incidents is inherently complex, as it requires engineers to interpret noisy telemetry signals, form hypotheses, and iteratively refine their understanding of the system under evolving context.

\textbf{Software Incident Timelines.} Our primary source of data comes from software incident timelines. These timelines document the real-time process of incident investigation and resolution, including Slack \cite{Slack2009} discussion threads between engineers at \CompanyName{} that begin when a software incident is detected and continue throughout the mitigation process. %
These timelines provide a real-time record of expert reasoning, including the questions engineers ask, the evidence they examine, and the conclusions they draw over time.

Interspersed with their natural language discussions, these engineers can also share embedded data visualization widgets in order to help triage and diagnose degraded systems. These visualizations encompass a wide range of telemetry data modalities, including logs, traces, and metrics. For \textsc{ARFBench}, we focus on metrics --- time series data that are a key tool for anomaly reasoning and cross-metric analysis in investigations.

\subsection{QA Generation Pipeline}
\label{sec: pipeline}
Our data collection process primarily involves three steps: data curation from real incidents, QA generation, and filtering and verification.

\textbf{``Oracle" Model.} In the construction of \textsc{ARFbench}, the discussion data is crucially used as extra context along with the time series plot as input to an ``oracle" VLM to generate choices and select putative labels for questions, akin to an offline version of how expert annotations are used to label anomalies in \citet{si2024timeseriesbench}. For particular question types, the incident timeline provides messages supporting exact answers within the text. Furthermore, having access to the incident timeline allows better understanding of the time series in comparison to other time series from the same incident. This allows for a zero-shot baseline that allows the model to ``cheat". 

We use the oracle VLM for the generation, filtering, and verification steps because it allows for automated, scalable question generation when analyzing complex incident data. The incident data often requires expert domain knowledge in specific software systems or infrastructure to understand, and having the VLM synthesize such information immensely speeds up this process to generate feasible answer choices. The answers are then verified by humans by using the incident report data, which allows us to prevent leakage of sensitive data and filter the questions for erroneous answer choices.

\textbf{Data Curation and Cleaning From Incidents. }  From each incident, we extract one or more time series to be used in QA generation. Each time series has an associated query string that maps to the given time series in a large-scale database. To remove sensitive information, we use an LLM to summarize and sanitize this query data, and we provide it as the time series caption for additional context at inference time. In addition, we include tag names corresponding to the different variates of the time series, giving context to what each channel represents (e.g. datacenter, asset, error type). See Section \ref{sec: ts-description} for further details on how the time series caption is generated.

\textbf{Answer Choice and Putative Label Generation. }  For each time series, we generate questions using a standardized set of templates, filling in relevant time series information such as the title and channel names. Following question generation, the question, time series plot, and full incident discussion data are used as context to prompt (App. \ref{app: option-generation-prompt}) the oracle VLM to generate answer choices relevant to the particular time series, as well as to choose a putative answer choice. Some question categories (i.e. Presence, Correlation, Indicator) have a fixed set of answer choices, with no VLM generation involved. For all other categories, the VLM generates anywhere between five and eight possible answer choices. 

\textbf{Filtering and Verification. } After generation, the questions are filtered by an LLM to ensure sensibility and relevance of the question-answer pairs, since some question types are more suitable for some time series than others, e.g. the question category is directly mentioned in the discussion of the incident timeline. Finally, the authors \textbf{manually filtered and corrected} all remaining QA pairs to ensure accuracy and no leakage of sensitive data. To ensure that each question category outside of Presence had exactly five answer choices, the correct answer and four randomly chosen answer choices are picked to be the final answer choice list.

\begin{table*}[!t]
\centering
\begin{minipage}[b]{0.29\textwidth}
\centering
\resizebox{\textwidth}{!}{%
\begin{tabular}{lc}
\toprule
Property & Number \\
\midrule
Questions & 750 \\
\# Distinct Timeseries & 142 \\
Data Points & 5.38M \\
\midrule
Tier I & 111 \\
Tier II & 306 \\
Tier III & 333 \\
\midrule
Min \# Variates & 1 \\
Median \# Variates & 10.5 \\
Maximum \# Variates & 2283 \\
\midrule
Median Length per Variate & 367 \\
Maximum Length per Variate & 40969\\
\bottomrule
\end{tabular}
}
\captionof{table}{Summary Statistics of \textsc{ARFBench}. ARFBench time series have high variate count and length, making input representation difficult for models.}
\label{tab:benchmark-statistics}
\end{minipage}
\hspace{1mm}
\hfill
\begin{minipage}[b]{0.66\textwidth}
\centering
\includegraphics[width=\linewidth]{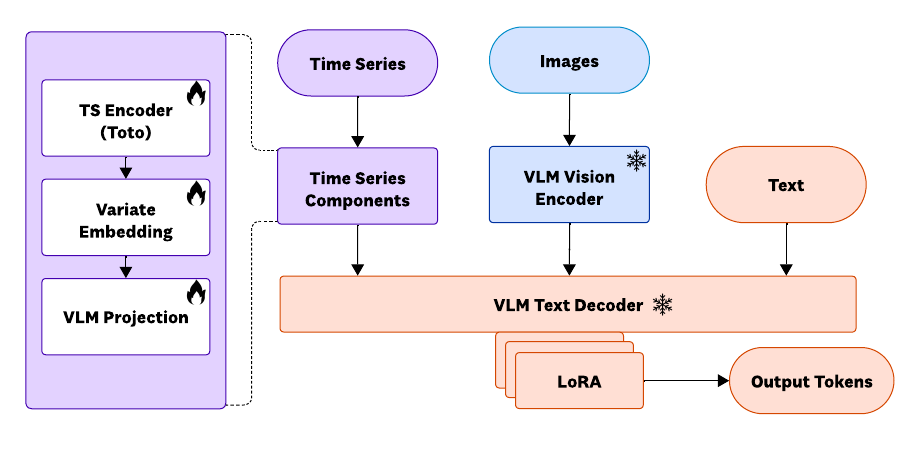}
\captionof{figure}{Architecture diagram of the Toto-1.0-QA-Experimental (Toto-Qwen3-VL) model. Frozen weights are denoted with a snowflake, while trainable weights are marked with a flame. Training a small set of weights allows us to adapt and augment the capabilities of either pretrained model.}
\label{fig:toto-vlm-arch}
\end{minipage}
\end{table*}
  
\section{Experimental Results}\label{sec:results}
\subsection{Evaluation Metrics}
\label{sec:metrics}
To evaluate each model, we compute overall, per-tier, and per-category accuracy and multiclass macro-F1. Multiclass macro-F1 is the unweighted mean of the F1 score computed separately for each class, treating each class in turn as the positive label and all others as negative. Accuracy and multiclass macro-F1 serve complementary roles as evaluation metrics. Accuracy measures overall task correctness and provides an interpretable measure of benchmark-level success, while macro-F1 measures whether that performance is balanced across answer classes rather than driven by a small subset of frequent labels.

\textbf{Macro-F1 prevents exploitation of accuracy. } The F1 score is used in conjunction with accuracy to account for the label imbalance inherent in incident data due to the high prevalence of anomalies. Baselines such as Frequent Choice, which always answer using the most common class, can yield deceptively high accuracy on some categories (e.g., Presence) while producing uninformative predictions. Multiclass macro-F1 explicitly penalizes such predictors, achieving at most $\frac{1}{\#\text{ classes}}$.

For each question, answer choices may be data-dependent (e.g. Magnitude: 1, 10, 100, 1000), so the raw answer choices are not comparable across questions.
Instead, we map raw answer choices to predefined sets of semantic answer classes for each category in a one-to-one mapping (e.g. Smallest, Small, Medium, Large, No Anomaly). The F1 score is computed across the semantic answer choices to be comparable across classes. The semantic class definitions are given in the Appendix \ref{app: semantic-bins}. We find that our per-category frequent choice baseline, which chooses the most frequent label in each category (and selects randomly if the most frequent label does not exist in the answer choices), yields the second lowest F1 of all models evaluated at 17.3\%.

\textbf{Accuracy complements macro-F1. } Macro-F1 is useful for preventing models from exploiting label imbalance, but it does not fully reflect performance on ARFBench. Since it weights all answer classes equally, macro-F1 can be sensitive to performance on rare classes and is less directly interpretable than accuracy. 

\subsection{Baselines and Model Details}
\label{sec:baselines}
\textbf{Models. } We use a few-shot prompting method (Appendix \ref{app: test-prompt}) for all models, regardless of post-training. We refer models without post-training as ``few shot". Among few-shot models, we evaluate 9 state-of-the-art open-source and proprietary VLMs on our benchmark, as well as 2 open-source and proprietary LLMs. Furthermore, we evaluate 2 existing post-trained foundation models, both of which are time series LLM models. For each model, we generally choose the newest available checkpoint in order to test the variant with the top expected reasoning capabilities. As a simple baseline, we evaluate Random Choice as well as per-category Frequent Choice, which picks the most frequently correct answer per category if it exists in the answer choices, otherwise a randomly selected valid answer is chosen. 

\textbf{User Study. } To compare against a human baseline, we conducted a user study with 4 time series researchers from \CompanyName{}: 2 observability domain experts and 2 non-domain experts. Users were asked to complete a randomly sampled 25\% set (188 QA pairs) of the benchmark, sampled from the same question category distribution as the full benchmark. 
Users were given a 16-question calibration set selected from the post-training dataset, which they were required to obtain at least 90\% accuracy to move on to the full user study.

\textbf{Model-Expert Oracle. } As a post-hoc analysis of our user study, we find that models and experts largely have different error distributions. Therefore, we posit a theoretical model-augmented expert oracle by taking the best-of-2 score between a domain expert and the Toto-1.0-QA-Experimental and having a perfect oracle judge pick the correct answer, if the correct answer is provided.  (Section \ref{sec: expert-vs-model}).

\subsection{Training Multimodal TSQA Models}
\label{sec:post-training}
\textbf{Motivation.} Time series in \textsc{ARFBench} are difficult to represent as input for few-shot LLMs and VLMs due to high variate count and length (Table \ref{tab:benchmark-statistics}). Furthermore, we find that existing few-shot models have major areas for improvement (Section \ref{sec:results}). This motivates the need for specialized time series foundation models. As an initial attempt in modeling, we post-trained three model families: VLMs, TSFM-LLMs, and TSFM-VLMs. Further details on post-training can be found in Appendix \ref{app: training}.

\textbf{\textsc{ARFBench} as an RLVR Task.} As defined in \citet{guan2025timeomni}, we pose \textsc{ARFBench} as a time series reinforcement learning with verifiable rewards (RLVR) task. In our setting, the model must output reasoning $\hat{R}$ and answer $\hat{y}$ given context (e.g. time series caption) $C$ and up to two time series inputs $X_i = \{x_{1:T_i}^{(v_i)}\}_{v_i=1}^{V_i}$, where $i \in \{1,2\}$ indexes the time series. $V_i$ and $T_i$ represent the number of channels and the length of time series $i$ respectively. A more formal setup is given in Appendix \ref{app: rl-formulation}.

\textbf{Reward Formulation.} Given ground-truth label $y$, we directly use outcome reward function $R(\hat y, y)\in \{0,1\}$ as an exact correctness reward based on our JSON parser. We do not give any reward for format or length, following practices in DAPO \cite{yu2025dapo}.

\textbf{Synthetic Training Data. } To post-train all models, we first generated 12,000 synthetic training examples by starting with random Gaussian noise and then injecting predefined anomalies (Figure \ref{fig:synthetic-data-gen}). The anomalies are injected uniformly at random into locations in the time range. Furthermore, the anomaly magnitude is scaled by a randomly chosen factor of the maximum value of the time series. 

\textbf{Real Training Data. } We also generated and manually labeled an additional set of 207 training examples, using the same pipeline as the main benchmark, but over a date range temporally disjoint and drawn from distinct incidents than the set of incidents used for \textsc{ARFBench}. The real labeled dataset was then augmented to 395 examples with negative tier III augmentation (Appendix \ref{app: tier-3-aug}).

\textbf{TSFM-VLM Architecture. } Training time series-vision-language models has only been recently explored for time series forecasting \cite{zhong2025time}. Such approaches rely on training from scratch. By contrast, recent work has shown promise in using pretrained TSFMs for time series classification \cite{auer2025pre}. Since \textsc{ARFBench} utilizes observability metrics, we jointly train a leading observability forecasting TSFM, Toto \cite{cohen2025time}, and a reasonably-sized leading open-source VLM, Qwen3-VL 32B \cite{bai2025qwen3vltechnicalreport} (Figure \ref{fig:toto-vlm-arch}). 

To adapt Toto and Qwen3-VL to work together, we train two components to transform embeddings from Toto into the VLM embedding space. First, the variate embedding multi-layer perceptron (MLP) aggregates embeddings across the time dimension, then normalizes and projects the resulting embeddings through a simple MLP. Importantly, the variate embedding MLP prevents the textual context length from scaling with the length of the series. The VLM projection layers then project the embeddings to the hidden dimension of the text decoder.

\textbf{Training Pipeline. } For both the TSFM-LLM and TSFM-VLM models, we used a three-stage training approach. The models were first trained with supervised fine-tuning (SFT) on the synthetic data (Stage 1), then with SFT on the real data (Stage 2). Stage 3 continues training from the Stage 2 checkpoint using RLVR with the DAPO recipe \cite{yu2025dapo}, a PPO-style asymmetric clipped objective. For the post-trained Qwen3-VL (without Toto), we omit the SFT phases and use only the RL stage, as this empirically led to the best results. For each stage, we trained a different LoRA \cite{hu2022lora} adapter for the text decoder and applied it sequentially before training the next stage. Early stopping was done for the SFT stages based on validation loss. We freeze prior-stage adapters when training subsequent stages; at evaluation we compose all adapters sequentially. Although this procedure is not standard continual post-training, it enables controlled isolation of stage-specific effects and supports mixing and matching adapters across stages. For the hybrid models, we additionally train projection layers as well as finetune the TSFM backbone. Hyperparameters are listed in Table \ref{tab: hyperparams}.

\subsection{Results}
\label{sec:main-results}
\begin{table*}[t]
    \centering
    \resizebox{\textwidth}{!}{%
    \begin{tabular}{lcccccccc}
    \toprule
    \multirow{2}{*}{\textbf{Model}} & \multirow{2}{*}{\textbf{Accuracy (95\% CI)} (\uparrow)} & \multicolumn{3}{c}{\textit{Accuracy by Tier}} & \multirow{2}{*}{\textbf{Overall F1 (95\% CI)} (\uparrow)} & \multicolumn{3}{c}{\textit{F1 by Tier}} \\
    \cmidrule(lr){3-5} \cmidrule(lr){7-9}
    & & \textit{Tier I} & \textit{Tier II} & \textit{Tier III} & & \textit{Tier I} & \textit{Tier II} & \textit{Tier III} \\
    \midrule
    \multicolumn{9}{c}{\textit{Baselines}} \\
    Random Choice & 24.5 & 50.0 & 20.0 & 20.0 & 22.5 & 45.6 & 20.5 & 16.8 \\
    Per-category Frequent Choice & 45.1 & 84.7 & 30.1 & 45.6 & 17.3 & 45.9 & 12.3 & 12.5 \\
    Non-domain Experts (n=2) & 69.7 & 80.4& 63.2& 72.0& 60.7 & 68.0 & 59.9 & 59.0 \\
    Domain Experts (n=2) & 72.7 & 89.3& 67.7& 71.4& 64.6& 76.1& 64.5& 60.9\\
    Model-Expert Oracle & 87.2 & 96.4& 80.3& 90.5& 82.8& 89.0& 77.1& 86.3\\
    \midrule
    \multicolumn{9}{c}{\textit{Large Language Models (LLMs)}} \\
    Qwen3 32B & 47.9 (44.41, 51.55)& 80.9& 35.1& 48.6& 36.1 (32.35, 39.14)& 55.7& 31.5& 33.8\\
    GPT-5 (text) & 56.4 (52.72, 60.05) & 82.6 & 45.2 & 57.9 & 43.9 (39.57, 47.24) & 66.1 & 39.6 & 40.3 \\
    \midrule
    \multicolumn{9}{c}{\textit{Vision-Language Models (VLMs)}} \\
    Qwen3-VL 8B & 45.3 (41.87, 49.07) & 80.2 & 40.8 & 37.8 & 34.7 (30.91, 37.88) & 63.5 & 36.1 & 23.6 \\
    Claude Sonnet 4.5 & 47.2 (43.86, 50.94) & 83.8 & 43.5 & 38.4 & 37.9 (34.08, 41.41) & 63.2 & 40.6 & 26.9 \\
    GPT-4o & 47.2 (43.60, 50.93) & 79.3 & 49.0 & 34.8 & 42.4 (37.11, 45.63) & 64.2 & 43.8 & 33.8 \\
    GPT-4.1 & 47.9 (44.27, 51.34) & 80.2 & 50.3 & 34.8 & 44.0 (38.49, 47.54) & 65.1 & 48.0 & 33.1 \\
    Qwen3-VL 32B & 52.8 (49.07, 56.67) & 80.2 & 46.7 & 49.2 & 45.1 (40.39, 48.77) & 65.1 & 41.9 & 41.3 \\
    Claude Opus 4.6 & 54.8 (51.47, 58.54) & \textbf{88.3} & 52.3 & 45.9 & 46.7 (41.45, 50.59) & 65.8 & 49.1 & 38.2 \\
    Gemini 3 Pro & 58.1 (54.80, 61.87) & 82.9 & 51.0 & 56.5 & 49.6 (44.85, 53.35) & \underline{67.8} & 49.7 & 43.4 \\
    GPT-5.4 & 61.3 (57.86, 64.94) & 81.1 & \underline{54.2} & \underline{61.3} & \underline{51.4} (46.57, 54.78) & 62.6 & \underline{50.4} & \textbf{48.4} \\
    GPT-5 & \underline{62.7} (59.20, 66.13) & 82.0 & \textbf{55.9} & \underline{62.5} & \textbf{51.9} (47.21, 55.38) & 66.9 & \textbf{51.2} & \underline{47.5} \\
    \midrule
    \multicolumn{9}{c}{\textit{Post-Trained Foundation Models}} \\
    OpenTSLM 1B (TS-LLM)& 0.8 (0.27, 1.60) & 0.0 & 2.0 & 0.0 & 1.2 (0.55, 1.84) & 0.0 & 3.0 & 0.0 \\
    ChatTS 8B (TS-LLM)& 31.1 (26.53, 32.80) & 60.4 & 26.5 & 25.5 & 22.1 (19.17, 24.71) & 48.1 & 20.0 & 15.4 \\
    Toto-1.0-Qwen3 32B (TSFM-LLM) & 48.8 (45.33, 52.27) & 82.9 & 47.4 & 38.7 & 33.9 (30.27, 36.77) & 60.0 & 43.6 & 16.4 \\
    Qwen3-VL 32B (post-trained) & 56.9 (53.47, 60.80) & \underline{84.7} & 50.3 & 53.8 & 46.6 (41.51, 50.61) & \textbf{69.8} & 44.9 & 40.5 \\
    Toto-1.0-QA-Experimental 32B (TSFM-VLM) & \textbf{63.9} (60.40, 67.07) & \underline{84.7} & \underline{55.6} & \textbf{64.6} & 48.9 (44.13, 52.27) & 66.3 & 48.4 & 43.5 \\
    \bottomrule
    \end{tabular}
    }
    \vspace{1mm}
    \caption{Accuracy and multiclass macro-F1 on ARFBench, sorted by model type and accuracy. Bootstrapped 95\% CIs resampled over 1000 batches of $n_{bootstrap}=750$ are provided for overall accuracy and F1. The best model score in each metric is bolded. The second best model score in each metric is underlined.}
    \label{tab:results-tiers-iclr}
    \end{table*} 
We present overall and per-tier accuracy and F1 scores for all models in Table \ref{tab:results-tiers-iclr}. Per-category results can be found in Table \ref{tab:results-category-acc-iclr} and \ref{tab:results-category-f1-iclr} in the Appendix.

\textbf{Frontier VLMs Substantially Lead Over Existing Baselines.} The top few-shot model on \textsc{ARFBench}, GPT-5, yielded a 62.7\% accuracy and 51.9\% F1. This improves significantly over random choice, at 24.5\% accuracy (22.5\% F1), as well as over other open-source and proprietary baselines outside the GPT-5 family by at least 4.6 percentage points (pp) accuracy and 2.3 pp F1. As expected, the models perform worse as the difficulty of the question increases. Tier III was the most difficult tier for all models to achieve high F1, with the best model achieving 48.4\% F1. This trend tends to persist in accuracy, though the top models tend to have higher accuracy in Tier III than Tier II. However, this can be contextualized by per-category frequent choice, showing that the label imbalance is greater in Tier III than in Tier II. 

We also corroborate several previously observed trends in TSQA. First, for both LLMs we evaluated, the overall and tier-wise performance lagged behind the VLM counterpart of the same model, aligning with results from \citet{daswani2024plots}.
Furthermore, while most VLMs are relatively strong at identifying the presence of an anomaly, they still struggle with reliably determining properties of the anomaly (Tier II) relative to humans, with a 7.3 pp accuracy and 13.3 pp F1 difference between non-domain experts and GPT-5. This matches the conclusions of \citet{zhou2024can}, particularly that visual perception in vision-language models does not match human perception of anomalies.

The top proprietary and open-source few-shot VLM models also showed a substantial gap in performance, with the best open-source model (Qwen3-VL 32B) performing 9.9 pp and 6.8 pp F1 worse overall compared to the best proprietary model (GPT-5). The main difference in performance between the two categories of models comes from Tiers II and III, where the model must reliably extract details from time series to answer the question. However, we find that some open-source models perform better than many older proprietary models or models from the Claude family.

\textbf{Hybrid TSFM-VLM Models Show Promise in TSQA.}  
Most time series LLMs feed time series as input to a time series encoder before passing the embedded input along to an LLM backbone. While these approaches may improve significantly over few-shot LLM-only inference of the base model, time series LLMs still lag significantly behind VLMs of similar size.

By contrast, our TSFM-VLM model, Toto-1.0-QA-Experimental, had comparable scores to top frontier models. Toto-1.0-QA-Experimental achieved the best accuracy (63.9\%) and fourth best F1 (48.9\% F1), surpassing the next best model (GPT-5) by 1.2 pp and lagging behind the best model in F1 (GPT-5) by 3 pp. Per tier, the Toto-1.0-QA-Experimental model yields the highest Tier III accuracy, as well as the second highest Tier 1 and Tier II accuracy. Toto-1.0-QA-Experimental also obtains comparably close tier-wise F1 scores to Gemini 3 Pro and improves significantly over the base model in all tiers, whereas other post-trained models improve in some tiers and degrade in others with respect to the base model. Analyzing the precision and recall of Toto-1.0-QA-Experimental across tiers, we find that the precision is consistently higher (2.8 pp to 15.5 pp) than the recall, suggesting that the model yields generally correct answers but under-predicts particular answer classes. 

As a preliminary experiment to compare the effect of data modalities in modeling, we compare post-trained Qwen3-VL 32B, Toto-1.0-Qwen3 32B, and Toto-1.0-QA-Experimental, which were all post-trained with the same training data, model size, and pipeline. Our results demonstrate that the time series and vision components of the architecture jointly play a large role in the performance gains of Toto-1.0-QA-Experimental, surpassing the other two models by at least 7 pp on accuracy and 2.3 pp F1 overall. 

\textbf{Model and Expert Performance Show Complementarity.}
\label{sec: expert-vs-model}
Given the 8.8 pp accuracy and 12.7 pp F1 gap in performance between the top model and expert performance, we wanted to understand whether models and users had largely similar errors, which would imply difficult or unanswerable questions. To compare, we took the results from the two top models by accuracy (GPT-5 and Toto-1.0-QA-Experimental) and the two domain experts across the sampled 25\% of questions used for the user study. We found that among questions that both experts got incorrect (23 questions), GPT-5 and Toto-1.0-QA-Experimental individually got 47.8\% (11 questions) and 34\% (8 questions) correct, respectively. Conversely, we find that among the questions that GPT-5 and Toto-1.0-QA-Experimental individually answer incorrectly (58 questions and 59 questions, respectively), at least one expert correctly answers the same questions 79.3\% (46 questions) and 74.6\% (44 questions) of the time. 
Therefore, given a large distributional mismatch between the questions models and humans answer correctly, we find evidence against unanswerable questions and an expert-level performance ceiling. 

Furthermore, we find that the errors made between models and experts are qualitatively different: experts tend to make fine-grained perceptual errors and instruction-following errors that may result from the lack of interactivity of the interface and tediousness of the task, while models tend to make domain-knowledge errors and hallucinate inconsistent answers at times. We present case studies in Appendix \ref{app:expert-comparison}.

To establish a baseline on the performance of experts when augmented with model predictions, we compute a best-of-2 score, where we assume a perfect judge that picks a correct answer whenever presented with two plausible choices. Using the results from GPT-5 and a domain expert, we hypothesize that this "Model-Expert Oracle" achieves 87.2\% accuracy and 82.8\% F1, demonstrating that under ideal conditions, experts augmented with model predictions perform far better than models or humans alone. \looseness=-1

\subsection{Ablations}
\label{app: ablations}

\begin{table*}[t]
\centering
\resizebox{\textwidth}{!}{%
\begin{tabular}{lcccccccc}
\toprule
\multirow{2}{*}{\textbf{Model}} & \multirow{2}{*}{\textbf{Accuracy} (\uparrow)} & \multicolumn{3}{c}{\textit{Accuracy by Tier}} & \multirow{2}{*}{\textbf{Overall F1} (\uparrow)} & \multicolumn{3}{c}{\textit{F1 by Tier}} \\
\cmidrule(lr){3-5} \cmidrule(lr){7-9}
& & \textit{Tier I} & \textit{Tier II} & \textit{Tier III} & & \textit{Tier I} & \textit{Tier II} & \textit{Tier III} \\
\midrule
Random Choice & 24.5 & 50.0 & 20.0 & 20.0 & 22.5 & 45.6 & 20.5 & 16.8 \\
Per-category Frequent Choice & 45.1 & 84.7 & 30.1 & 45.6 & 17.3 & 45.9 & 12.3 & 12.5 \\
\midrule
GPT-4o (No TS)& 35.2& \textbf{82.9}& 22.9& 30.6&  20.0& 16.5& 13.2& 20.0\\
GPT-4o (No Caption)&  \textbf{49.3}& 79.3& 47.7& \textbf{40.8}& 42.0& 61.0& 42.3& \textbf{35.4}\\
GPT-4o (Full Input) & 47.2 & 79.3 & \textbf{49.0}& 34.8 & \textbf{42.4}& \textbf{64.2}& \textbf{43.8}& 33.8 \\
\midrule
GPT-5 (No TS)& 32.9& \textbf{82.9}& 21.6& 26.7&  22.3& 53.9& 14.5& 18.8\\
GPT-5 (No Caption) &  56.3& 80.2& 49.7& 54.4& 46.5& 64.0& 46.8& 40.4\\
GPT-5 (Full Input)& \textbf{62.7}& 82.0 & \textbf{55.9}& \textbf{62.5}& \textbf{51.9}& \textbf{66.9}& \textbf{51.2} & \textbf{47.5}\\

\bottomrule
\end{tabular}
}
\vspace{1mm}
\caption{GPT-4o and GPT-5 accuracy and F1 on ARFBench when various parts of the input are ablated. No TS data means the timeseries image was not given to the model (i.e. the answers are only dependent on the time series description). No Caption means the time series description was not given to the model (i.e. the model only has the time series plot as input). Full input means both the description and plot were present.  The best performing ablation for each model (excluding baselines) is highlighted for each question category.}
\label{tab:ablations-tiers-iclr}
\end{table*} \textbf{Ablating Training Stages of Toto-1.0-QA-Experimental.} As discussed in Section \ref{sec:post-training}, Toto-1.0-QA-Experimental is trained in three separate stages: an SFT stage with synthetic data, an SFT stage with real data, and a real data RLVR stage. We find that the performance at each stage varies heavily. The performance after the initial SFT stage is poor: a near 0 score is obtained across most questions. Since the model generates reasoning blocks without a final answer, it yields no score for a majority of questions. After the second SFT stage, Toto-1.0-QA-Experimental yields a 48.4\% accuracy and 37.8\% F1, which is significantly lower than the base Qwen3-VL 32B model performance (52.8\% accuracy and 45.1\% F1). However, this stage has relatively few answer formatting issues, suggesting that the VLM has adapted to utilizing the time series tokens. The final RLVR stage refines the model's ability to use time series embeddings and raises the accuracy by 15.4 pp (11 pp F1) from the second SFT stage. 

By contrast, we also find that the SFT stages play an important role in the joint adaptation of the TSFM and VLM. By omitting one or both SFT stages, we found that the RL training was significantly slower to improve in the first 100 steps, achieving a per-step mean reward that was around 2-4 times less than the per-step mean reward after the two SFT stages. We leave the question of understanding the role of the SFT stages in RL as future work to understanding training dynamics for TSFM-VLM models.

\textbf{VLM Context Ablations.} We experiment with context ablations to test the significance of each component of the multimodal input context in model performance (Table \ref{tab:ablations-tiers-iclr}). To do this, we test GPT-4o and GPT-5 on \textsc{ARFBench} when there are (1) no provided plots of the time series (2) no time series descriptions (captions) to condition on what the time series represents.
We find that, for both models, when no time series data is provided (in image or text format), the model performs between random choice and per-category frequent choice in overall and per-tier F1 as well as accuracy.  When only the time series data is provided but not the caption, GPT-5 performs substantially better (+23.4 pp accuracy / +24.2 pp F1) than GPT-5 with no time series data, but still fairly below GPT-5 with full input. By contrast, we find that GPT-4o performs better in accuracy when no caption is provided, though overall F1 still remains worse than having full input. Thus, additional context is not uniformly utilized across models in ARFBench.  
\section{Limitations and Future Work}
\textsc{ARFBench} presents a novel benchmark for evaluating the capabilities of multimodal models and introduces new modeling approaches to TSQA. However, \textsc{ARFBench} leaves out more open-ended questions commonly asked in the incident response process, such as common mitigation strategies (e.g. \textit{Based on this anomaly, should we scale up resources?}) or linking to common incident patterns (e.g. \textit{Is this anomaly directly linked to a deployment?}), which may require further labeled data.
In addition, \textsc{ARFBench} is still limited to single-turn QA, which means it is not yet a comprehensive benchmark for agentic tasks or QA. On the modeling end, there is more work to understand why TSFM-VLM models perform so well on certain tasks and how to improve such models. We leave such tasks as future work. 
 
\section{Conclusion}
In this work, we present \textsc{ARFBench}, a TSQA benchmark designed to test reasoning capabilities grounded in the software observability domain. We demonstrate that novel joint time series and vision-language modeling shows great promise, and we hope that the open-source release of \textsc{ARFBench} will be a valuable resource for the ML community to build upon existing methods for time series and software incident response.  
\section*{Acknowledgements}
We thank Afshin Rostamizadeh, Zongzhe Xu, Mohamed El Yamani, Wayne Chi, Misha Khodak, Samuel Müller, Gerald Woo, Valerie Chen, Anna Wei, Shanda Li for feedback and helpful comments on the paper. We also thank Chris Lettieri, Amine Naouas, Max Livingston for help in the data collection process, and Qiqi Ren for help in testing various components of the benchmark.

This material is based upon work supported by the National Science Foundation Graduate Research Fellowship Program under Grant No(s) DGE2140739. Any opinions, findings, and conclusions or recommendations expressed in this material are those of the
author(s) and do not necessarily reflect the views of the National Science Foundation. 
\bibliography{refs}

\newpage
\appendix
\onecolumn
\section{Additional Details on Benchmark Construction Pipeline}
\label{app: benchmark-composition}
All questions from the benchmark are generated from 63 incidents resolved on March 7th, 2025 to March 30th, 2025. Each time series within each incident is extracted, the channel names are anonymized, and a plot is generated for the time series. Due to computational overhead, we limit the time series we extract to under 2500 non-empty variates. For each time series, all single time series question templates are applied to generate a candidate question. For each incident, we select random pairs of time series (up to 10 pairs) and apply all paired time series question templates. For each question, we also prompt an LLM to select a putative answer. Then, all questions are filtered according to our filtering prompt, which aims to remove questions that clearly do not have overlaps in time range. The remaining questions are then uploaded to the open source labeling platform Label Studio, where the authors reviewed and changed the putative answers if necessary.

\subsection{Time Series Caption}
\label{sec: ts-description}
\begin{figure}[h]
    \centering
    \includegraphics[width=\linewidth]{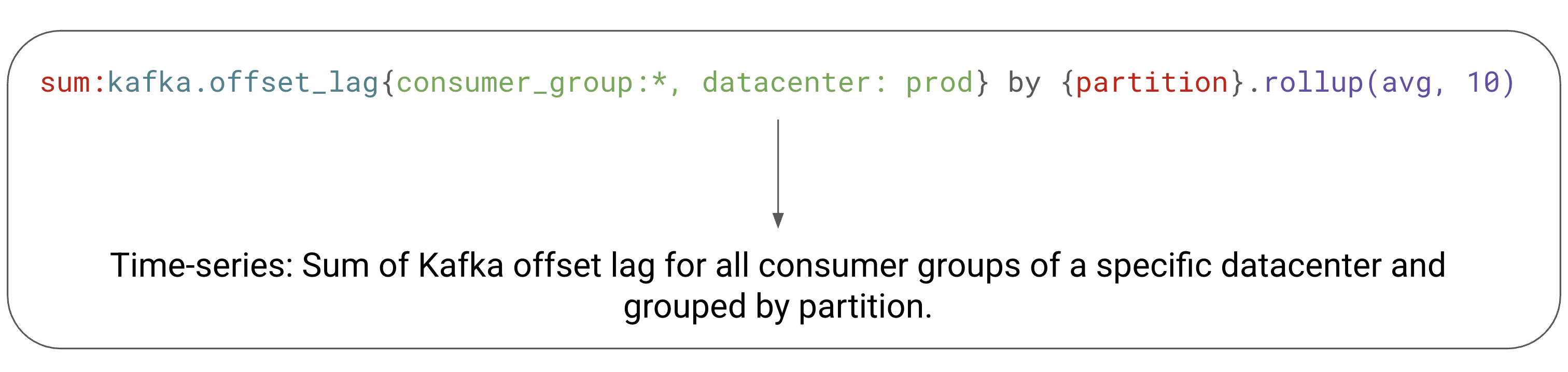}
    \caption{\textbf{Top:} Example metric query used to query the time series database. The blue text represents the metric name. The green text represents the filters. The red text represents space aggregation functions and group-by keys. Finally, the purple text represents a time aggregation function. \textbf{Bottom:} Example summarized time series description of the metric query above.}
    \label{fig:query_summary}
\end{figure}

The time series description or caption comes from summarizing database queries used to fetch the corresponding time series (Figure \ref{fig:query_summary}). This database query language utilizes space aggregations, time aggregations, filters, and more to select particular time series variates from a large database. In the construction of \textsc{ARFBench}, we only collect query strings from internal users and systems and do not use any customer data. These query strings are then summarized with an LLM to anonymize sensitive information and used as context in each question, as explained in Section \ref{sec: pipeline}. 

\subsection{Software Incident Timelines In Context of Incident Response}
As mentioned, software incident timelines provide a source of expert annotations to help label our benchmark. Connecting to broader incident response workflow, the incident timeline logs the overall process of incident response, which typically consists of (but is not limited to) following nonlinear steps:
\begin{enumerate}[wide, labelwidth=!, labelindent=0pt]
    \item \textbf{Incident Report:} Alerting or triggering the beginning of the incident.
    \item \textbf{Diagnosis:} Understanding the severity, systems, and customers that are affected by the incident.
    \item \textbf{Triage:} Prioritizing the workflow to target different impacted components based on severity and other factors.
    \item \textbf{Mitigation:} Working towards changes that reduce the negative effects of the incident.
    \item \textbf{Resolution:} Finding solutions that resolve the downstream impacts of the incident.
    \item \textbf{Root Cause Analysis (RCA):} Finding the specific cause of the incident.
    \item \textbf{Postmortem:} Creating a write-up of what lessons were learned from the incident.
\end{enumerate}

TSQA and reasoning can be a crucial tool in many of these steps, especially in triage, mitigation, and RCA. For example, looking at a time series of error rates can help to triage by identifying the portion of traffic affected. Comparing multiple time series can help identify which system component is the source of errors, aiding in RCA. Moreover, after a mitigation is attempted, responders may want to assess whether anomalous behavior in a time series has resolved to a normal state.

\subsection{Observability Time Series Subdomains}
\begin{figure*}[t]
  \centering
  \begin{minipage}[t]{0.49\textwidth}
    \centering
    \includegraphics[width=\linewidth]{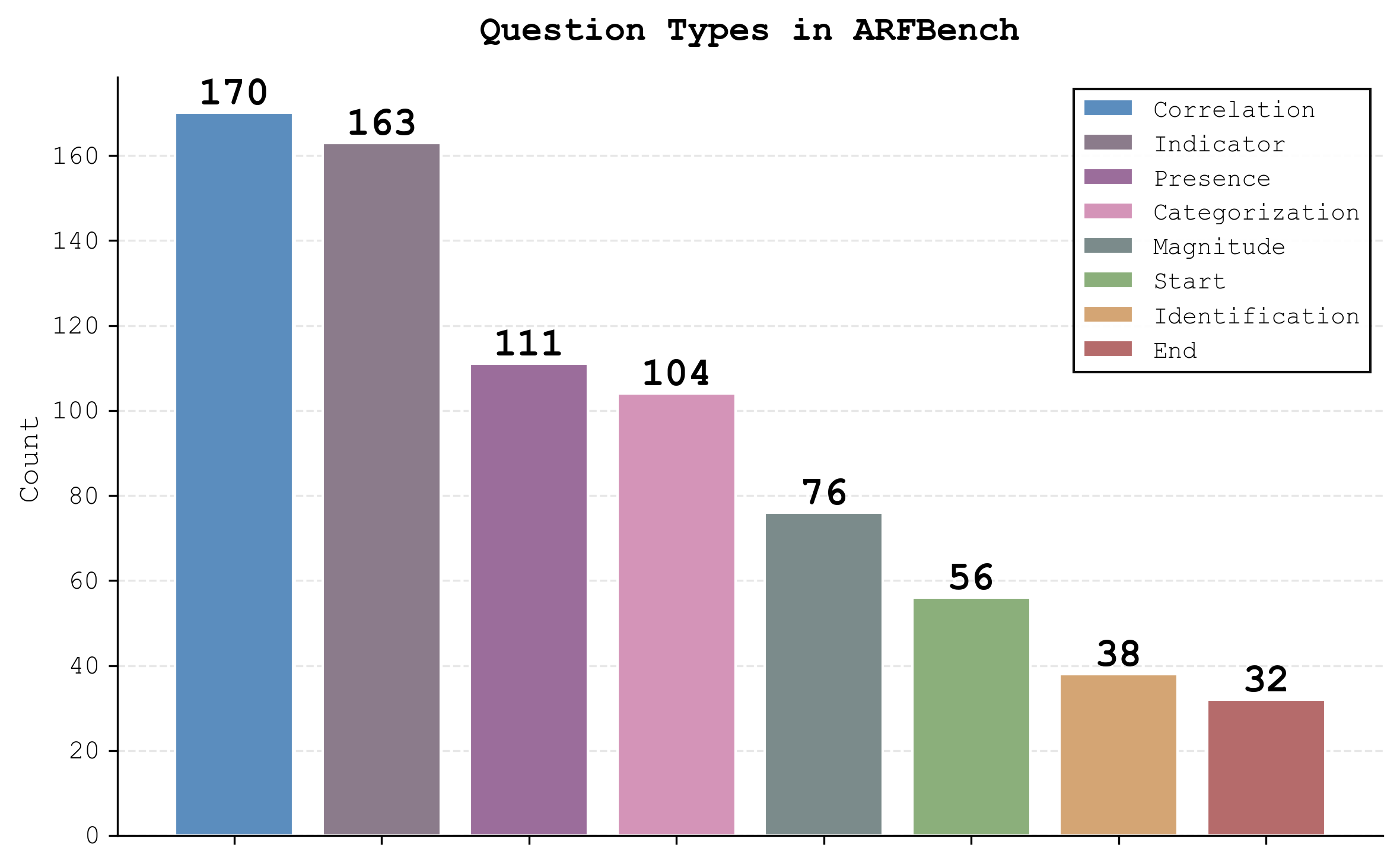}
  \end{minipage}\hfill
  \begin{minipage}[t]{0.49\textwidth}
    \centering
    \includegraphics[width=\linewidth]{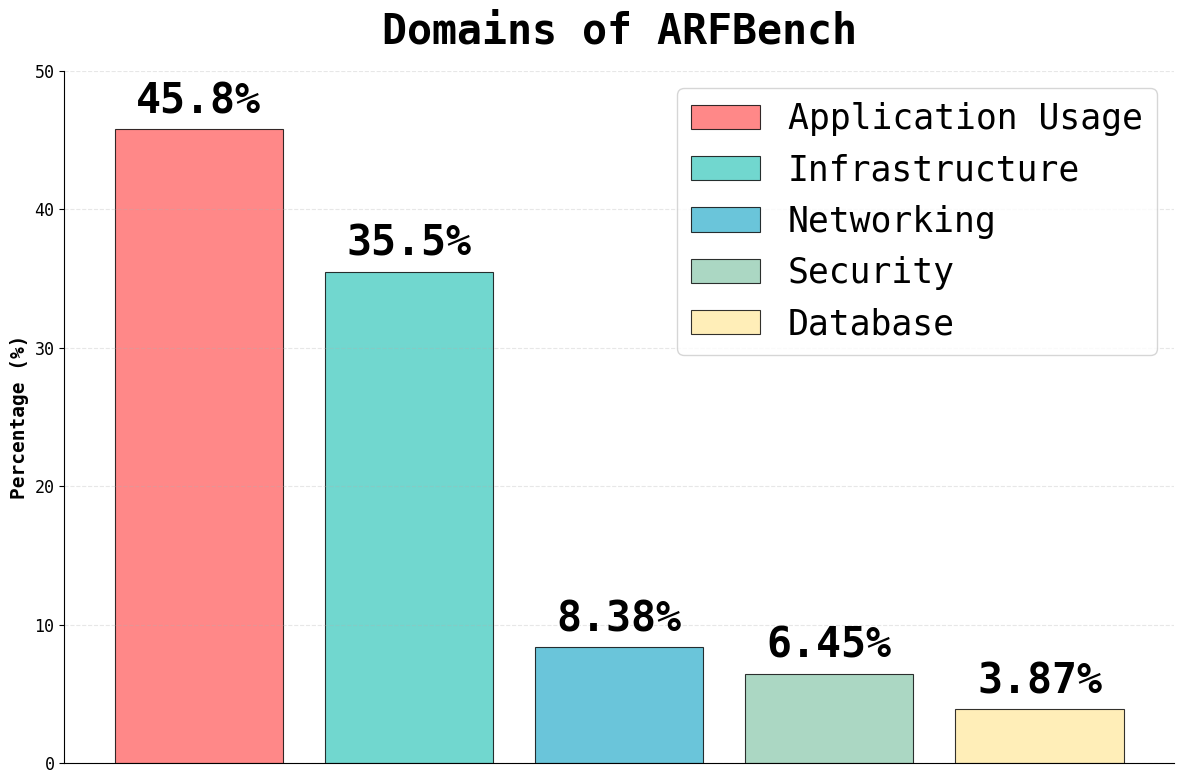}
  \end{minipage}

  \caption{Distribution of question categories (left) and time series domains (right) in ARFBench.}
  \label{fig: categories-domains}
\end{figure*}

Using the captions of each time series, we classify each time series into one or more of various subdomains of observability (Figure \ref{fig: categories-domains}). These include:
\begin{enumerate}[wide, labelwidth=!, labelindent=0pt]
    \item \textbf{Application Usage}: data collected from the application layer within a service
    \item \textbf{Infrastructure}: metrics of common system measurements, such as CPU usage or GPU utilization
    \item \textbf{Networking}: metrics of network- and communication-related measurements, e.g. latency
    \item \textbf{Database}: data collected when performing database operations, such as querying
    \item \textbf{Security}: metrics that monitor the general security of a system, such as vulnerability messages
\end{enumerate}
Including each of these domains ensures a diverse benchmark with time series of diverse characteristics. To categorize the domains of the time series in \textsc{ARFBench} we use GPT-4.1 to classify each query string.

\subsection{Answer Choice Filtering}
During the benchmark construction phase, the LLM pipeline will potentially generate more answer choices than are used in the final QA pair, depending on the question category. This is reflected in the prompts. We postprocess the benchmark to filter down to five answer choices per category, following the semantic answer classes given in Appendix \ref{app: details}.

\subsection{Negative Tier III Augmentation}
\label{app: tier-3-aug}
During question generation, questions in Tier III are generated between pairs of time series from the same incident. This often results in correlated time series, leading to a biased label distribution that may be unrepresentative of the task at hand in the real-world (i.e. the pairs of time series may not always be hand-selected by experts when a model is asked to compare them). In order to smooth the biased label distribution in Tier III after the data pipeline, we augment Tier III of the benchmark with questions between time series that are not in the same incident, which we determine to not be correlated or indicative of each other.

\section{Details on Question Categories}
\label{app: details}
\begin{table}[!t]
\resizebox{\textwidth}{!}{%
\begin{tabular}{>{\centering\arraybackslash}p{2.5cm}   %
  >{\centering\arraybackslash}p{4cm}     %
  >{\centering\arraybackslash}p{2.5cm}   %
  p{6.5cm}
}
\toprule
\textbf{Time Series Scope} & \textbf{Question Category}  &\textbf{Difficulty}& \textbf{Example Template Question} \\
\midrule
\multirow{16}{*}{\textbf{Single}} 
 & Presence  &Tier I& \texttt{Does the time series exhibit an anomaly in the given time range?}\\
\cmidrule(lr){2-4}
 & Identification  &Tier II& \texttt{Which channels are exhibiting anomalies in the given time range?} \\
\cmidrule(lr){2-4}
 & Start Time  &Tier II& \texttt{What is the start time of the anomaly, if an anomaly exists?} \\
\cmidrule(lr){2-4}
 & End Time  &Tier II& \texttt{What is the end time of the anomaly, if an anomaly exists?} \\
\cmidrule(lr){2-4}
 & Magnitude  &Tier II& \texttt{How much does the anomaly in this time-series deviate from the expected behavior of this time-series, if an anomaly exists?} \\
\cmidrule(lr){2-4}
 & Categorization  &Tier II& \texttt{What type of anomaly in the given time range is exhibited, if any?} \\
\midrule
\multirow{8}{*}{\textbf{Paired}} 
 & Correlation  &Tier III& \texttt{Does the anomaly in time-series 1 correlate with the anomaly in the other time-series, if anomalies exist?} \\

\cmidrule(lr){2-4}
 & Leading/Lagging Indicator  &Tier III& \texttt{Is the anomaly in time-series 1 a leading or lagging indicator of the anomaly in time-series 2, if anomalies exist?} \\
\bottomrule
\end{tabular}
}
\vspace{1mm}
\caption{An overview of the question categories within ARFBench. Questions are ordered by increasing difficulty, corresponding to the required reasoning and synthesis across all given inputs. Each increasing tier of difficulty requires the correct reasoning of the previous lower tier. }
\label{tab: question_categories}
\end{table} In the following section, we provide details on the construction of each question category and associated answer choices. In all questions, an anomaly is not guaranteed to exist, and an option for ``No Anomaly'' is always provided. We also sort each category's answer choices into two or five predefined semantic answer choices to keep consistent binning for multiclass F1 score.
\subsection{Question Definitions}
\label{app: question-defs}
\textbf{Presence.} The Presence question is designed to test whether a model can distinguish the existence of an anomaly in the given time series. An anomaly exists in a time series if the time series includes a subsequence of points that an expert engineer would take action upon to investigate.
A time series is not limited to one anomaly, however, the question is designed to test for at least one anomaly.
The answer choices for the presence question are in binary Yes/No format. 

\textbf{Identification. }The Identification question tests whether a model can distinguish which subset of channels within the given three channels within a time series are anomalous. The given channels are chosen randomly. In total, there are five answer choices, two answer choices of single variate answer choices, one pair of variates, and the triplet of all three variates in question.

\textbf{Start Time. }The Start Time question tests whether a model can find the time stamp at which an anomaly begins. An anomaly may not exist within the time series. Within a given time series, the start time refers to the beginning of the first anomaly of the time series. The anomaly may begin before the start of the time series data and this case should be jointly inferred from the time series description and data. Answer choices correspond to different time stamps along the time series, with one answer as Before the earliest time stamp of the given options. 

\textbf{End Time. }The End Time question tests whether a model can find the time stamp at which an anomaly ends. Within a given time series, the end time refers to the end of the last anomaly of the time series. The anomaly may end after the end of the time series data and this case should be jointly inferred from the time series description and data. Answer choices correspond to different time stamps along the time series, which are not guaranteed to lie within the time range of the time series, with one answer as Not resolved referring to the end time of the anomaly being after the end of the given time series data. 

\textbf{Magnitude. }The Magnitude question tests whether a model can correctly assess the deviation of the time series from an expected counterfactual forecast. In general, this is calculated as the maximum ratio between the anomalous values and the mean counterfactual values. In the case that the counterfactual values are zero, the magnitude is instead calculated as the maximum absolute value of the anomalous values. For this question, the LLM is prompted to generate reasonable answer choices in a logarithmic scale. 

\textbf{Categorization.} The Categorization question tests whether a model can correctly classify an anomaly into one of five anomaly categories relevant for software incident response, which include: level shift, transient spike, change in seasonality, change in variance, and change in trend. We qualitatively describe these categories in relation to the traditional categories of anomalies in anomaly detection \cite{chandola2009anomaly} (i.e. point, contextual, and collective anomalies), but we find that the granularity of the definitions for anomaly detection is not sufficient for analyzing anomalies relevant in software incident response.

A level shift represents a prolonged change in mean value of the time series. The level shift may arise as a contextual anomaly, where the mean value is substantially different than what is expected from the general value of the time series in relation to its time series description, or it may be a collective anomaly, in which we see an identifiable subsequence of the data have a change in mean value over time. 

A transient spike is a type of point anomaly, where a small collection of points are substantially different in magnitude than other points in the time series. Transient spikes often return to the original values and are very short in duration, leading to little to no shift in mean value.

A change in seasonality is a type of contextual or collective anomaly, where the magnitude of the time series does not change, but the seasonal pattern of the time series changes in frequency. 

A change in variance describes when the variance of the time series significantly shifts from its previous value for a sustained period of time.

A change in trend is a contextual or collective anomaly in which the time series increases or decreases for a significant period of time. 

Multiple such anomalies may occur in one series, but in general, we consider the one with the largest magnitude or most relevant to the incident timeline to be the correct answer. 

\textbf{Correlation.} The Correlation question tests whether a model correctly identifies correlations between two time series as well as a model can correctly identify the existence of anomalies in both. Two anomalies are correlated if they have temporally similar patterns in context to the time series descriptions, or if they have known relation through the software infrastructure based on what they represent.

\textbf{Indicator. }The Indicator question is a follow-up question to the correlation question, testing a model on whether it can identify the temporal order between anomalies. The answer choices include that the first anomaly is a leading indicator of the second anomaly, a lagging indicator of the second anomaly, perfectly correlated between the two anomalies, or not correlated at all.

\subsection{Semantic Binning Categories}
\label{app: semantic-bins}
\textbf{Presence. }The semantic classes of Presence are the same as the answer choice classes (i.e. Yes/No).

\textbf{Identification. }The semantic classes of Identification are No Anomaly (in the channels listed), One Channel (small), One Channel (large), Two Channels, Three Channels, with the single variate answer choices ordered lexicographically.

\textbf{Start Time. }The semantic classes of Start Time are No Anomaly, Earliest, Early, Medium, Late, corresponding to the temporal order of the answer choices presented.

\textbf{End Time. }The semantic classes of End Time are No Anomaly, Early, Medium, Late, Latest, corresponding to the temporal order of the answer choices presented.

\textbf{Magnitude. }The semantic classes of Magnitude are No Anomaly, Smallest, Small, Medium, Large, corresponding to the magnitude order of the answer choices presented.

\textbf{Categorization. }The semantic classes for Categorization are No Anomaly, Change in Trend, Transient Spike, Level Shift, Change in Seasonality/Variance.

\textbf{Correlation. }The semantic classes match the answer choices exactly. These include: No, there is no anomaly in either time-series, No, there is an anomaly in both but they are not correlated, No, there is an anomaly only in time-series 1, No, there is an anomaly only in time-series 2, Yes, there is an anomaly in both and they are correlated.

\textbf{Indicator. }The semantic classes match the answer choices exactly. These include: The anomaly in time-series 1 is a leading indicator of the anomaly in time-series 2, The anomaly in time-series 1 is a lagging indicator of the anomaly in time-series 2, The anomaly in time-series 1 is not correlated to the anomaly in time-series 2, The anomaly in time-series 1 is perfectly correlated to the anomaly in time-series 2, No Anomaly in one or both series.

\section{Additional Details on LLM Settings for Inference}

\textbf{Inference.} To evaluate each of the foundation models, we used one of two methods. First, closed-source models were accessed through their respective APIs: Claude Sonnet 4.5 and Claude Opus 4.6 was evaluated through the Anthropic API \cite{Claude4.5Sonnet2025}, GPT-5, GPT-5.4, GPT-4.1, and GPT-4o were evaluated with the OpenAI API \cite{OpenAIGPT4}, and Gemini 3 Pro Preview was evaluated through the Google Gemini API.

Open-weight models including Qwen3-VL 8B, 32B, Qwen3 32B
were evaluated via the vLLM library \cite{kwon2023efficient}, while models such as ChatTS, OpenTSLM, Toto-1.0-Qwen3, Toto-1.0-QA-Experimental were evaluted with Transformers.  All models were evaluated with eight NVIDIA A100 GPUs with tensor parallel, using weights downloaded from Hugging Face Transformers \cite{wolf2019huggingface}.

In terms of model choice, we preferred newer models due to higher overall accuracy on multimodal benchmarks. For the OpenTSLM series, we choose the soft-prompt (SP) Llama-3.2 1B model due to its high test performance on benchmarks reported in the OpenTSLM paper \cite{langer2025opentslm}.

For inference, we use temperature 0.05 for all third-party models for factuality and determinism (except GPT-5, GPT-5.4, Gemini 3 Pro, which had default temperature 1.0 and is unconfigurable by the API) with maximum token output of 2000 tokens. For reasoning models, we set their reasoning level to ``medium" with 2000 tokens. We tested using ``high" with 10000 tokens and found similar results to the ones reported, however, we include only the 2000 token results for fairness of comparison with other models. In cases where reasoning models gave blank answers due to reaching or surpassing the maximum output tokens, we labeled such answers as incorrect. For the Toto-1.0-QA-Experimental and Toto-1.0-Qwen3 models, we use temperature 0.

Due to context length and channel quantity limits on ChatTS, we make several preprocessing decisions before inputting the time series to the model. To process time series longer than 1024, we subsample the time series down to 1024. For time series with more than 50 variates, we select the 50 variates with the highest mean from each time series in the case of single time series questions, and the 25 variates with the highest mean from each time series in the case of paired time series questions. For the GPT-5 text and Qwen3 32B models, which have a longer context length, we first find a single optimal interval for which the entire series can fit in the context length, and if not, we instead truncate a fraction of the time series data from the middle of the series at various truncation ratios (25\%, 40\%, 55\%, 70\%) until we achieve the desired context.

\textbf{Multiple Choice Shuffling. } At inference time, we randomly shuffle the answer choices before building the prompt in order to prevent option order bias that may result from any one instantiation of ARFBench. 

\textbf{Plot Generation.} In order to evaluate vision-language models with time series data, we transform each time series into a plot with Matplotlib (for single series) and Plotnine (for paired series) with 100 dots per inch (DPI) and then downsize each plot to have a maximum size of 1500 pixels on each side in order to satisfy API requirements. For Qwen3-VL we allow models to auto-resize the images using the built-in Qwen utility tools. For Tier I and Tier II images, the model receives one image as input, corresponding to the time series. We only include channel names in the plot for Identification questions. In Tier III, we input three images: the first image has both time series stacked vertically with a shared time axis. The channels are monochromatic, i.e. all channels of the first time series are blue and all channels of the second time series are orange. The last two plots are plots of each individual time series, in the order they are referenced in the question. 

Furthermore, the channel names of the time series are omitted from the time series plot in order to retain image resolution and clarity, unless there are less than 8 channels. The channel names are provided in other components (e.g. prompt or question) of the textual input for all QA pairs.

\section{Additional Details on Training}
\label{app: training}

We trained three types of models: vision-language models, TSFM-LLM models, and TSFM-VLM models. 

\subsection{Data}
Both synthetic data and real data were used for post-training. Synthetic data was generated by generating scaled random Gaussian noise time series of various lengths, from 240 to 50,000 timestamps, with 40\% of series distributed within 1000-10,000 timestamps. The time series are randomly perturbed with seasonal and drift components for diversity. We predefine several anomaly types, including level shifts, transient spikes, changes in variance, seasonality, and flatline anomalies. For question generation, we generate an equal number of each category, randomly selecting an anomaly (or no anomaly at all) to inject into the synthetic time series. The timestamp of anomaly injection is also uniformly chosen across the length of the time series. We generated a total of 12,000 synthetic QA pairs for our training set.

The real data was generated with the ARFBench pipeline on incidents resolved on April 1st to April 7th, 2025. The incidents were manually cleaned to ensure no data overlap with the benchmark data. A total of 207 examples were labeled, which was then augmented to 395 examples via Tier III augmentation. 

For both sets of training data, synthetic reasoning traces were added by prompting a VLM for a reasoning explanation given the question and correct answer. These traces were used as supervision for both SFT stage 1 and stage 2.
\begin{figure}
    \centering
    \includegraphics[width=0.75\linewidth]{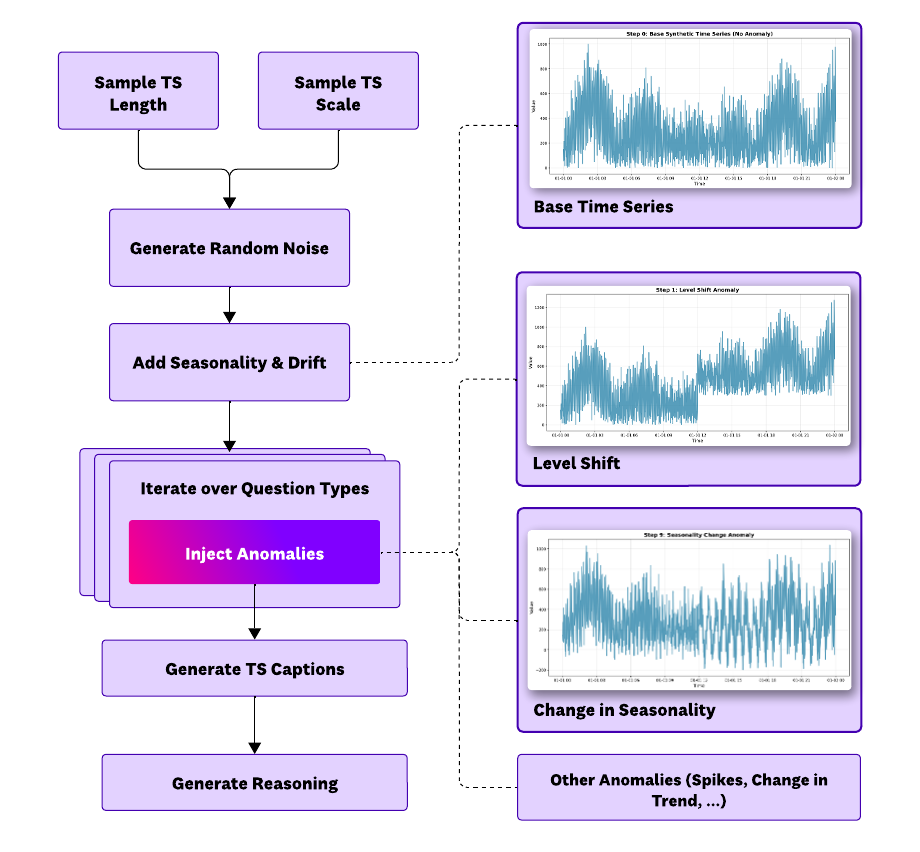}
    \caption{Process of generating synthetic multivariate data for post-training VLMs, TSFM-VLMs, and TSFM-LLMs. Gaussian noise is transformed by adding seasonality and drift components, then anomalies are injected at randomly chosen points within the time series. To make the data more realistic, captions, channel names, and reasoning are generated with an LLM to reflect the observability setting.}
    \label{fig:synthetic-data-gen}
\end{figure}
\subsection{Training Frameworks and Hardware}
VLM post-training was done with the veRL \cite{sheng2025hybridflow} framework with 8 NVIDIA H200s, with training and inference done in the FP16 precision \cite{qi2025defeating}. Toto-1.0-Qwen3 and Toto-1.0-QA-Experimental models were trained with TRL and patched to support time series input. Training was done with 8 NVIDIA B200s for Toto-1.0-Qwen3 and Toto-1.0-QA-Experimental. For Toto related models, some additional data was provided in the prompt for context: the start and end absolute timestamp, the timestep interval, and variate names in the order they were provided to Toto. 

\subsection{RL Formulation.}
\label{app: rl-formulation}
Recall our time series RLVR setting: for context (including prompt, question, and additional context) $C$ and time series input $\{X_j\}_{j=1}^{k}$ for $k\in \{1,2\}$, we sample a group $G$ of outputs $\{o_i\}_{i=1}^{n} \sim \pi_{\theta_{\text{old}}}(\cdot \mid C, \{X_j\}_{j=1}^{k})$, where the output $o_i = (\hat R_i, \hat y_i)$ includes both reasoning and model answer. For each group, each output $o_i$ receives a reward $r_i$ as defined by our 0-1 reward function in Section \ref{sec:post-training}. 

DAPO \cite{yu2025dapo} optimizes our multimodal models $\pi_\theta$ with the following objective:
\[
\mathcal{J}(\theta)
= \frac{1}{n}\sum_{i=1}^n \frac{1}{|o_i|}
\sum_{q=1}^{|o_i|}
\min\!\left(
\phi_{i,q}(\theta)\,\hat A_i,\;
\text{clip}\!\left(\phi_{i,q}(\theta),
1-\epsilon_{\text{low}}, 1+\epsilon_{\text{hi}}\right)\hat A_i
\right),
\]
where $\phi_{i,q}(\theta)$ is the token-level importance ratio
\[
\phi_{i,q}(\theta)
= \frac{\pi_\theta(o_{i,q} \mid C, \{X_j\}_{j=1}^{k}, o_{i,<q})}
{\pi_{\theta_{\text{old}}}(o_{i,q} \mid C, \{X_j\}_{j=1}^{k}, o_{i,<q})},
\]
and $\hat A_i$ is the group-relative advantage computed from rewards:
\[
\hat A_i =
\frac{r_i - \text{mean}(\{r_i\}_{i=1}^n)}
{\text{std}(\{r_i\}_{i=1}^n)}.
\]

\begin{table}[t]
\centering
\small
\setlength{\tabcolsep}{5pt}
\begin{tabular}{lccc}
\toprule
\textbf{Hyperparameter} & \textbf{VLM} & \textbf{TSFM--LLM} & \textbf{TSFM--VLM} \\
\midrule
\multicolumn{4}{l}{\textit{Common}} \\
Optimizer                            & \texttt{AdamW} & \texttt{AdamW} & \texttt{AdamW} \\
Precision (training \& inference)                    & \texttt{fp16} & \texttt{bf16} & \texttt{bf16} \\
Number of devices                    & 8 & 8 & 8 \\
\midrule
\multicolumn{4}{l}{\textit{LoRA / adapters}} \\
Trainable parameters                 & \texttt{LoRA only} & \texttt{LoRA + Toto + proj}& \texttt{LoRA + Toto + proj}\\
Freeze vision tower                  & \texttt{True} & \texttt{True} & \texttt{True} \\
Freeze TS Encoder                    & -- & \texttt{False} & \texttt{False} \\
\midrule
\multicolumn{4}{l}{\textit{SFT Stage 1 (synthetic)}} \\
Used?                                & \texttt{No} & \texttt{Yes} & \texttt{Yes} \\
Epochs                               & -- & \texttt{3} & \texttt{3} \\
Learning rate                        & -- & \texttt{1e-05} & \texttt{1e-05} \\
Batch size (global)                  & -- & \texttt{8} & \texttt{8} \\
Warmup steps                         & -- & \texttt{10} & \texttt{10} \\
Weight decay                         & -- & \texttt{1e-2} & \texttt{1e-2} \\
Max time series length               & -- & \texttt{51200} & \texttt{51200} \\
Max variates                         & -- & \texttt{100} & \texttt{100} \\
\midrule
\multicolumn{4}{l}{\textit{SFT Stage 2 (real)}} \\
Used?                                & \texttt{No} & \texttt{Yes} & \texttt{Yes} \\
Epochs                               & --& \texttt{4} & \texttt{4} \\
Learning rate                        & --& \texttt{1e-05} & \texttt{1e-05} \\
Global Batch size                 & --& \texttt{8} & \texttt{8} \\
Warmup steps                         & --& \texttt{10} & \texttt{10} \\
Weight decay                         & --& \texttt{1e-2} & \texttt{1e-2} \\
Max time series length               & -- & \texttt{25600} & \texttt{25600} \\
Max variates                         & -- & \texttt{100} & \texttt{100} \\
\midrule
\multicolumn{4}{l}{\textit{RL Stage 3 (DAPO)}} \\
Max context length                   & \texttt{20000} & \texttt{20000} & \texttt{20000} \\
Global batch size                      & \texttt{16} & \texttt{40} & \texttt{32} \\
Generations per prompt ($n$)            & \texttt{8} & \texttt{8} & \texttt{8} \\
$\epsilon_{low}$ (DAPO low clip coefficient)   & \texttt{0.2} & \texttt{0.2} & \texttt{0.2} \\
$\epsilon_{hi}$ (DAPO high clip coefficient) & \texttt{0.28} & \texttt{0.28} & \texttt{0.28} \\
Policy learning rate                 & \texttt{1e-06} & \texttt{2e-05}& \texttt{1e-05}\\
Max time series length               & -- & \texttt{25600} & \texttt{25600} \\
Max variates                         & -- & \texttt{100} & \texttt{100} \\
Temperature                         & -- & \texttt{1.3} & \texttt{0.8}\\
\bottomrule
\end{tabular}
\caption{Post-training hyperparameters for SFT and RL (DAPO) across model families.}
\label{tab: hyperparams}
\end{table}

\section{Additional Results}
\begin{table*}[t]
\centering
\resizebox{\textwidth}{!}{%
\begin{tabular}{lccccccccc}
\toprule
\multirow{3}{*}{\textbf{Model}} & \multirow{3}{*}{\textbf{Overall Accuracy} (\uparrow)} & \multicolumn{8}{c}{\textit{Accuracy by Category}} \\
\cmidrule(lr){3-10}
& & \multicolumn{1}{c}{\textit{Tier I}} & \multicolumn{5}{c}{\textit{Tier II}} & \multicolumn{2}{c}{\textit{Tier III}} \\
\cmidrule(lr){3-3} \cmidrule(lr){4-8} \cmidrule(lr){9-10}
& & \textbf{Presence} & \textbf{Identification} & \textbf{Start Time} & \textbf{End Time} & \textbf{Magnitude} & \textbf{Categorization} & \textbf{Correlation} & \textbf{Indicator} \\
\midrule
\multicolumn{10}{c}{\textit{Baselines}} \\
Random Choice & 24.5 & 50.0 & 20.0 & 20.0 & 20.0 & 20.0 & 20.0 & 20.0 & 20.0 \\
Per-category Frequent Choice & 45.1 & 84.7 & 36.8 & 35.7 & 34.4 & 17.1 & 32.7 & 42.9 & 48.5 \\
Non-domain Experts (n=2) & 69.7& 80.4& 66.7& 64.3& 68.8& 60.5& 61.5& 72.1& 72.0\\
Domain Experts (n=2) & 72.7& 89.3& 77.8& 67.9& 75& 60.5& 72.4& 74.4& 68.3\\
Model-Expert Oracle & 87.2& 96.4& 77.8& 78.6& 100.0 & 68.4& 84.6& 95.4& 85.4\\
\midrule
\multicolumn{10}{c}{\textit{Large Language Models (LLMs)}} \\
Qwen3 32B (text) & 47.9& 80.9& 28.9& 27.3& 35.5& 37.3& 39.8& 50.9& 46.3\\
GPT-5 (text) & 56.4 & 82.6 & \underline{47.4} & 29.6 & 38.7 & 51.4 & 50.0 & 56.9 & 59.0 \\
\midrule
\multicolumn{10}{c}{\textit{Vision-Language Models (VLMs)}} \\
Qwen3-VL 8B & 45.3 & 80.2 & 26.3 & 25.0 & 31.3 & 57.9 & 45.2 & 57.1 & 17.8 \\
Claude Sonnet 4.5 & 47.2 & 83.8 & 18.4 & 30.4 & 37.5 & 53.9 & 53.8 & 58.8 & 17.2 \\
GPT-4o & 47.2 & 79.3 & 39.5 & 35.7 & 43.8 & 61.8 & 51.9 & 45.3 & 23.9 \\
GPT-4.1 & 47.9 & 80.2 & 28.9 & 33.9 & 40.6 & \textbf{68.4} & 56.7 & 45.9 & 23.3 \\
Qwen3-VL 32B & 52.8 & 80.2 & 23.7 & 33.9 & 56.3 & 59.2 & 50.0 & 61.8 & 36.2 \\
Claude Opus 4.6 & 54.8 & \textbf{88.3} & 31.6 & 37.5 & 53.1 & 57.9 & 63.5 & 65.9 & 25.2 \\
Gemini 3 Pro & 58.1 & 82.9 & 28.9 & \underline{44.6} & 62.5 & 56.7 & 54.8 & \textbf{71.2} & 41.1 \\
GPT-5.4 & 61.3 & 81.1 & 31.6 & \textbf{63.6} & \underline{65.6} & 57.9 & 56.7 & 61.8 & \underline{60.7} \\
GPT-5 & \underline{62.7} & 82.0 & 31.6 & \underline{44.6} & \textbf{68.8} & \underline{65.8} & 59.6 & 63.5 & \textbf{61.3} \\
\midrule
\multicolumn{10}{c}{\textit{Post-Trained Foundation Models}} \\
OpenTSLM (TS-LLM) & 0.8 & 0.0 & 0.0 & 3.6 & 0.0 & 5.3 & 0.0 & 0.0 & 0.0 \\
ChatTS (TS-LLM) & 31.1 & 59.5 & 15.8 & 16.1 & 15.6 & 28.9 & 20.2 & 40.0 & 14.7 \\
Toto-1.0-Qwen3 (TSFM-LLM) & 48.8 & 82.9 & 10.5 & 35.7 & 34.4 & 47.4 & \textbf{71.2} & 41.8 & 35.6 \\
Qwen3-VL 32B (post-trained) & 56.9 & \underline{84.7} & 36.8 & 41.1 & 43.8 & 63.2 & 52.9 & \underline{67.6} & 39.3 \\
Toto-1.0-QA-Experimental (TSFM-VLM) & \textbf{63.9} & \underline{84.7} & \textbf{47.4} & 26.8 & 59.4 & \underline{64.5} & \underline{66.3} & \underline{68.8} & \underline{60.1} \\
\bottomrule
\end{tabular}
}
\vspace{1mm}
\caption{Accuracy on ARFBench across question categories and models. For Random Choice only, invalid answers are randomly shuffled to incorrect answer classes. The best model performance in each metric is bolded. The second best model performance in each metric is underlined. Our model, Toto-1.0-QA-Experimental, achieves the top accuracy among models tested.}
\label{tab:results-category-acc-iclr}
\end{table*} %
\begin{table*}[t]
\centering
\resizebox{\textwidth}{!}{%
\begin{tabular}{lccccccccc}
\toprule
\multirow{3}{*}{\textbf{Model}} & \multirow{3}{*}{\textbf{Overall F1} (\uparrow)} & \multicolumn{8}{c}{\textit{F1 by Category}} \\
\cmidrule(lr){3-10}
& & \multicolumn{1}{c}{\textit{Tier I}} & \multicolumn{5}{c}{\textit{Tier II}} & \multicolumn{2}{c}{\textit{Tier III}} \\
\cmidrule(lr){3-3} \cmidrule(lr){4-8} \cmidrule(lr){9-10}
& & \textbf{Presence} & \textbf{Identification} & \textbf{Start Time} & \textbf{End Time} & \textbf{Magnitude} & \textbf{Categorization} & \textbf{Correlation} & \textbf{Indicator} \\
\midrule
\multicolumn{10}{c}{\textit{Baselines}} \\
Random Choice & 22.5 & 45.6 & 21.2 & 18.9 & 18.2 & 20.4 & 21.7 & 15.8 & 17.8 \\
Per-category Frequent Choice & 17.3 & 45.9 & 10.8 & 16.3 & 14.1 & 6.0 & 14.6 & 12.0 & 13.1 \\
Non-domain Experts (n=2) & 61.3& 68.0 & 79.0& 67.4& 67.2& 40.3& 61.2& 58.4& 62.4\\
Domain Experts (n=2) & 64.6& 76.1& 77.5& 74.2& 72.6& 51.8& 67.3& 64.1& 57.6\\
Model-Expert Oracle & 82.8& 89.0& 68.3& 83.4& 100.0& 67.0& 75.6& 94.4& 77.8\\
\midrule
\multicolumn{10}{c}{\textit{Large Language Models (LLMs)}} \\
Qwen3 32B& 36.1& 55.7& 28.4& 26.6& 26.9& 31.4& 36.8& 32.3& 35.4\\
GPT-5 (text) & 43.9& 66.1 & 38.1 & 27.9 & 27.0 & 44.8 & 47.6 & 38.0 & 42.4 \\
\midrule
\multicolumn{10}{c}{\textit{Vision-Language Models (VLMs)}} \\
Qwen3-VL 8B & 34.7 & 63.5 & 28.6 & 21.8 & 23.5 & 47.0 & 42.8 & 33.1 & 13.8 \\
Claude Sonnet 4.5 & 37.9 & 63.2 & 16.8 & 33.2 & 31.3 & 49.3 & 49.8 & 33.8 & 19.8 \\
GPT-4o & 42.4 & 64.2 & 34.6 & 30.3 & 36.1 & 51.8 & 50.8 & 40.1 & 27.2 \\
GPT-4.1 & 44.0 & 65.1 & 29.2 & 33.5 & 32.7 & \textbf{63.7} & 55.9 & 42.9 & 23.3 \\
Qwen3-VL 32B & 45.1 & 65.1 & 25.0 & 30.8 & 46.7 & 46.9 & 49.0 & 47.5 & 34.7 \\
Claude Opus 4.6 & 46.7 & 65.8 & 34.3 & 36.1 & 45.1 & 53.8 & \underline{59.2} & \underline{51.6} & 24.1 \\
Gemini 3 Pro & 49.6 & \underline{67.8} & 38.6 & 43.3 & \textbf{57.1} & 50.3 & 54.5 & \textbf{57.0} & 29.2 \\
GPT-5.4 & \underline{51.4} & 62.6 & 29.6 & \textbf{53.3} & \underline{55.1} & 51.7 & 54.1 & 47.7 & \textbf{49.1} \\
GPT-5 & \textbf{51.9} & 66.8 & 32.8 & \underline{44.2} & 47.8 & \underline{59.1} & 57.0 & 49.0 & \underline{45.9} \\
\midrule
\multicolumn{10}{c}{\textit{Post-Trained Foundation Models}} \\
OpenTSLM 1B (TS-LLM) & 1.2 & 0.0 & 8.2 & 2.7 & 0.0 & 6.0 & 0.0 & 0.0 & 0.0 \\
ChatTS 8B (TS-LLM) & 22.1 & 48.1 & 22.2 & 15.0 & 14.4 & 27.9 & 17.9 & 21.4 & 9.2 \\
Toto-1.0-Qwen3 32B (TSFM-LLM)& 33.9 & 59.9 & 17.5 & 41.3 & 23.0 & 35.9 & \textbf{66.2} & 18.6 & 14.1 \\
Qwen3-VL 32B (post-trained) & 46.6 & \textbf{69.7} & \underline{40.5} & 37.2 & 36.7 & 48.9 & 50.3 & 46.8 & 33.9 \\
Toto-1.0-QA-Experimental 32B (TSFM-VLM)& 48.9 & 66.3 & \textbf{46.9} & 23.0 & 48.8 & 54.1 & 58.4 & 44.2 & 42.7 \\
\bottomrule
\end{tabular}
}
\vspace{1mm}
\caption{Multiclass F1 on ARFBench across question categories and models. Invalid answers are randomly shuffled to incorrect answer classes and results are averaged over 10 random seeds where applicable. The best model performance in each metric is bolded. The second best model performance in each metric is underlined. }
\label{tab:results-category-f1-iclr}
\end{table*} \subsection{Per Category Results}
The per-category accuracy and F1 results are shown in Tables \ref{tab:results-category-acc-iclr} and \ref{tab:results-category-f1-iclr}, respectively.

\section{Error Analysis and Case Studies}
\label{app:case-studies}
\label{sec:error-analysis}
We qualitatively analyze the errors made by the models on the benchmark to further understand model output. To do this, we have the model output a reasoning explanation. We sample 50 such errors across all models and categorize them into three different error categories: incorrect perception, instruction following errors, and limited context usage errors.

\textbf{Incorrect Perception (48\%).} These errors result from incorrect analysis of the time series. This includes hallucinating time series values or time ranges, identifying the wrong anomaly time range, or reasoning about incorrect details regarding the time series itself. 

\textbf{Limited Context Usage (42\%).} These errors result when not all provided information is utilized. For example, models may neglect processing all channels, leading to information loss. Without utilizing all channels, the model may not be able to contextualize the anomaly properly. In general, this can include incorrect multivariate understanding, lack of caption usage, or lack of domain knowledge.

\textbf{Instruction Following Errors (10\%).} These errors occur from incorrect instruction following. For example, in many Correlation questions, models answer ``The anomaly in time series 1 is not correlated to the anomaly in time series 2" even after deducing that there is no anomaly one of the time series.

\begin{figure}[hp]
\centering

\begin{outercasebox}{Case Study: Limited Context Usage}

\begin{casebox}[title={Question}, colback=black!2]{Question}
In the following time-series, what is the start time of the anomaly, if an anomaly exists?

\textbf{Time-series:} Sum of Kafka partition time lag for ingestion, filtered by consumer group and datacenter, and grouped by datacenter, topic, and partition.
\end{casebox}

\begin{tcolorbox}[colback=black!2,colframe=black!2,boxrule=0pt,
                  left=0pt,right=0pt,top=0pt,bottom=0pt,
                  height=0.28\textheight,valign=center]
\centering
\includegraphics[height=\tcbtextheight,width=\linewidth,keepaspectratio]{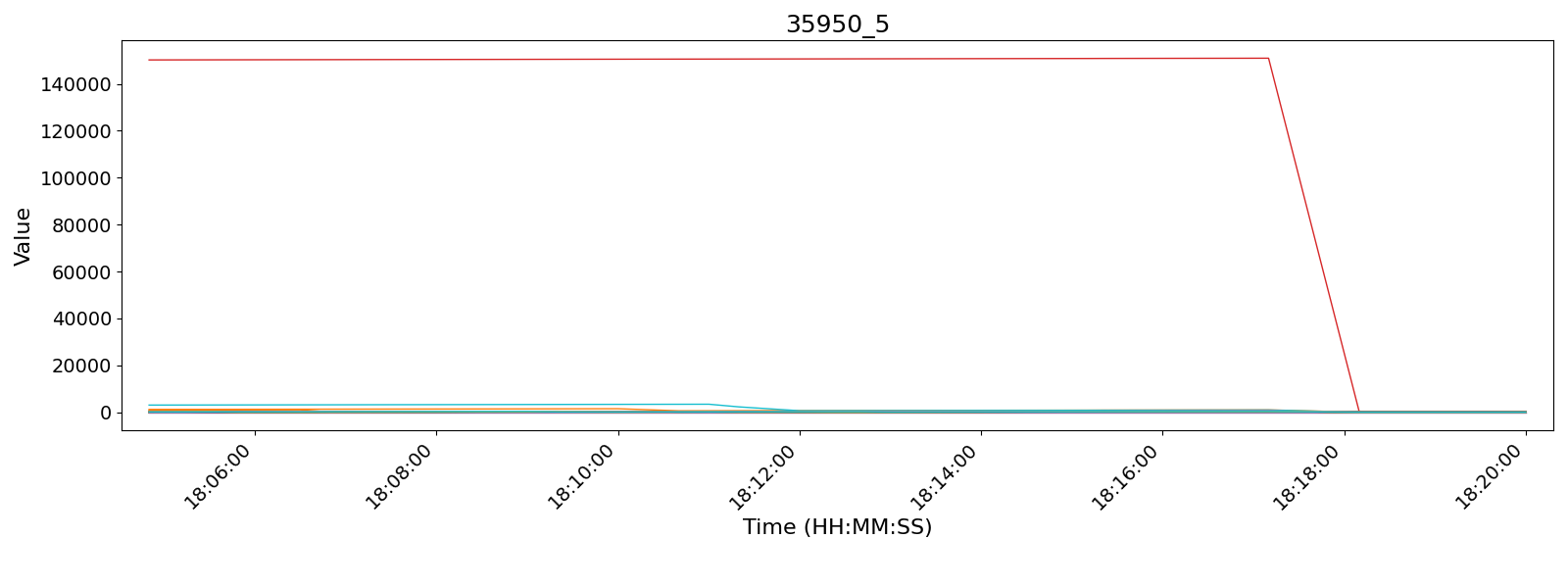}
\end{tcolorbox}

\begin{tcolorbox}[colback=white,colframe=white,boxrule=0pt,arc=0pt,
                  left=0pt,right=0pt,top=0pt,bottom=0pt] 
\begin{tcbraster}[
  raster columns=2,
  raster column skip=8pt,
  raster left skip=0pt,
  raster right skip=0pt,
  raster equal height=rows,
  raster width=\linewidth
]
\begin{answerbox}{red}
\textbf{Model answer:} 2025-03-07 18:15:00
\end{answerbox}
\begin{answerbox}{green}
\textbf{Correct answer:} Before the earliest timestamp
\end{answerbox}
\end{tcbraster}
\end{tcolorbox}

\begin{casebox}[title={Model reasoning}, colback=red!6, colframe=red!40]{Model reasoning}
\textbf{ChatTS:} The time series shows a downward spike with an amplitude of about 150913.00 between point 38 and point 45. The start time of this spike is around point 38, which corresponds to the timestamp 2025-03-07 18:15:00.
\end{casebox}

\begin{casebox}[title={Correct reasoning}, colback=green!6, colframe=green!40]{Correct reasoning}
Kafka lag is unusually high for the red channel, as we can see by comparing to other channels present in the graph that are close to 0, even before the time series begins. This denotes that the anomaly began before the window given in the time series.
\end{casebox}

\end{outercasebox}

\caption{An example where ChatTS does not utilize the extra textual context given in the question to solve the problem.}
\label{fig:ea-2}

\end{figure}

\begin{figure}[hp]
\centering

\begin{outercasebox}{Case Study: Domain Knowledge / Limited Context Usage}

\begin{casebox}[title={Question}, colback=black!2]{Question}
In the following time-series, what type of anomaly in the given time range is exhibited, if any?

Time-series: Average of container I/O write and read operations for a specific PostgreSQL service in a targeted environment, grouped by pod and device, with maximum rollup applied.
\end{casebox}

\begin{tcolorbox}[colback=black!2,colframe=black!2,boxrule=0pt,
                  left=0pt,right=0pt,top=0pt,bottom=0pt,
                  height=0.28\textheight,valign=center]
\centering
\includegraphics[height=\tcbtextheight,width=\linewidth,keepaspectratio]{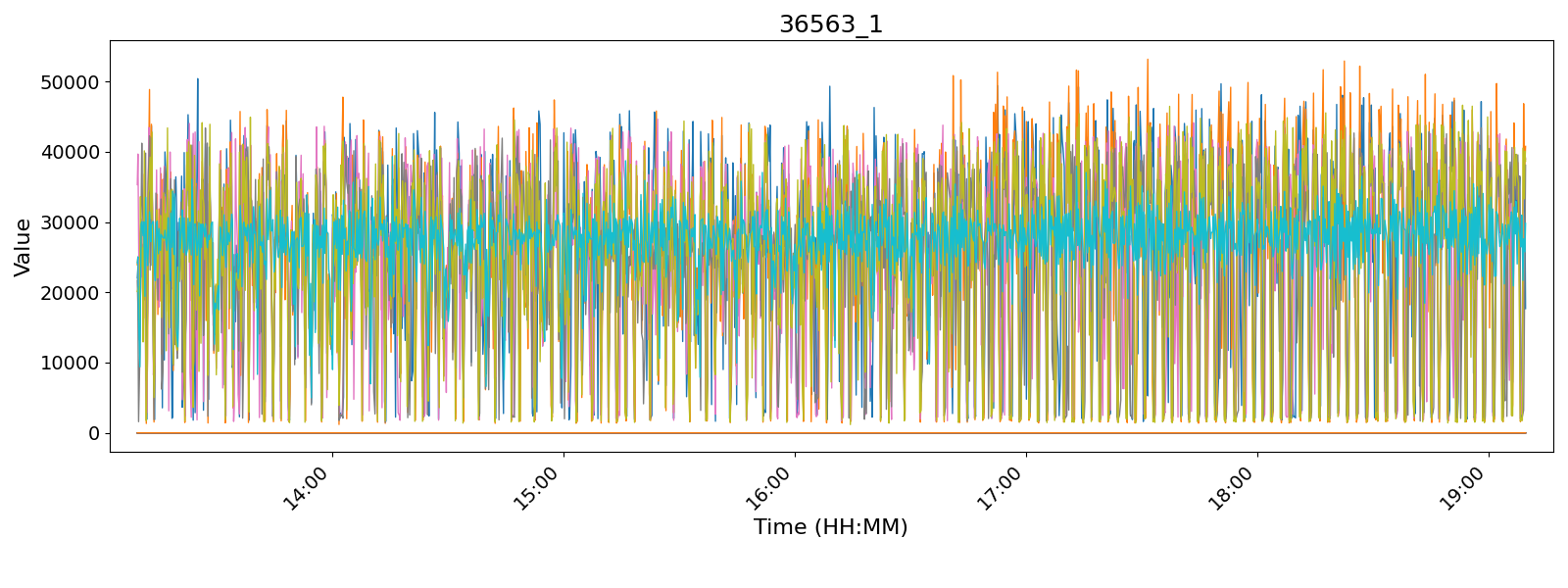}
\end{tcolorbox}

\begin{tcolorbox}[colback=white,colframe=white,boxrule=0pt,arc=0pt,
                  left=0pt,right=0pt,top=0pt,bottom=0pt] 
\begin{tcbraster}[
  raster columns=2,
  raster column skip=8pt,
  raster left skip=0pt,
  raster right skip=0pt,
  raster equal height=rows,
  raster width=\linewidth
]
\begin{answerbox}{red}
\textbf{Model answer:} No Anomaly
\end{answerbox}
\begin{answerbox}{green}
\textbf{Correct answer:} Level Shift
\end{answerbox}
\end{tcbraster}
\end{tcolorbox}

\begin{casebox}[title={Model reasoning}, colback=red!6, colframe=red!40]{Model reasoning}
\textbf{GPT-4.1:} The time-series plot shows multiple channels with high-frequency fluctuations, but there is no visible sustained change in mean (level shift), no sudden spike that returns to normal (transient spike), no obvious change in the periodic pattern (seasonality), and no clear change in the overall trend. The data appears to be consistently noisy throughout the observed period, with no significant deviation from the established pattern. Therefore, there is no evidence of an anomaly in the given time range.
\end{casebox}

\begin{casebox}[title={Correct reasoning}, colback=green!6, colframe=green!40]{Correct reasoning}
The time series begins to shift upwards and spike (e.g. the gold/orange channel) much more often after around 16:30 which is anomalous for the container I/O. For this question, we note that human annotations and time series models (i.e. Toto-1.0-QA-Experimental) are able to perceive this level shift, though it can be difficult to visualize on a plot without sufficient training.
\end{casebox}

\end{outercasebox}

\caption{An example where GPT-4.1 (VLM) lacks the domain knowledge and perceptual ability to notice an anomaly. }
\label{fig:ea-3}

\end{figure}

\begin{figure}[hp]
\centering

\begin{outercasebox}{Case Study: Domain Knowledge}

\begin{casebox}[title={Question}, colback=black!2]{Question}
Is the anomaly in time-series 1 a leading or lagging indicator of the anomaly in time-series 2, if anomalies exist?
\begin{itemize}
    \item \textbf{Time-series 1}: Average utilization percentage for a specific metadata queue in a targeted datacenter, grouped by Kubernetes deployment and datacenter.
    \item \textbf{Time-series 2:} Sum of available replicas in Kubernetes deployments for a specific service, filtered by datacenter and grouped by deployment and datacenter.
\end{itemize}
\end{casebox}

\begin{tcolorbox}[colback=black!2,colframe=black!2,boxrule=0pt,
                  left=0pt,right=0pt,top=0pt,bottom=0pt,
                  height=0.42\textheight,valign=center]
\centering
\includegraphics[height=\tcbtextheight,width=\linewidth,keepaspectratio]{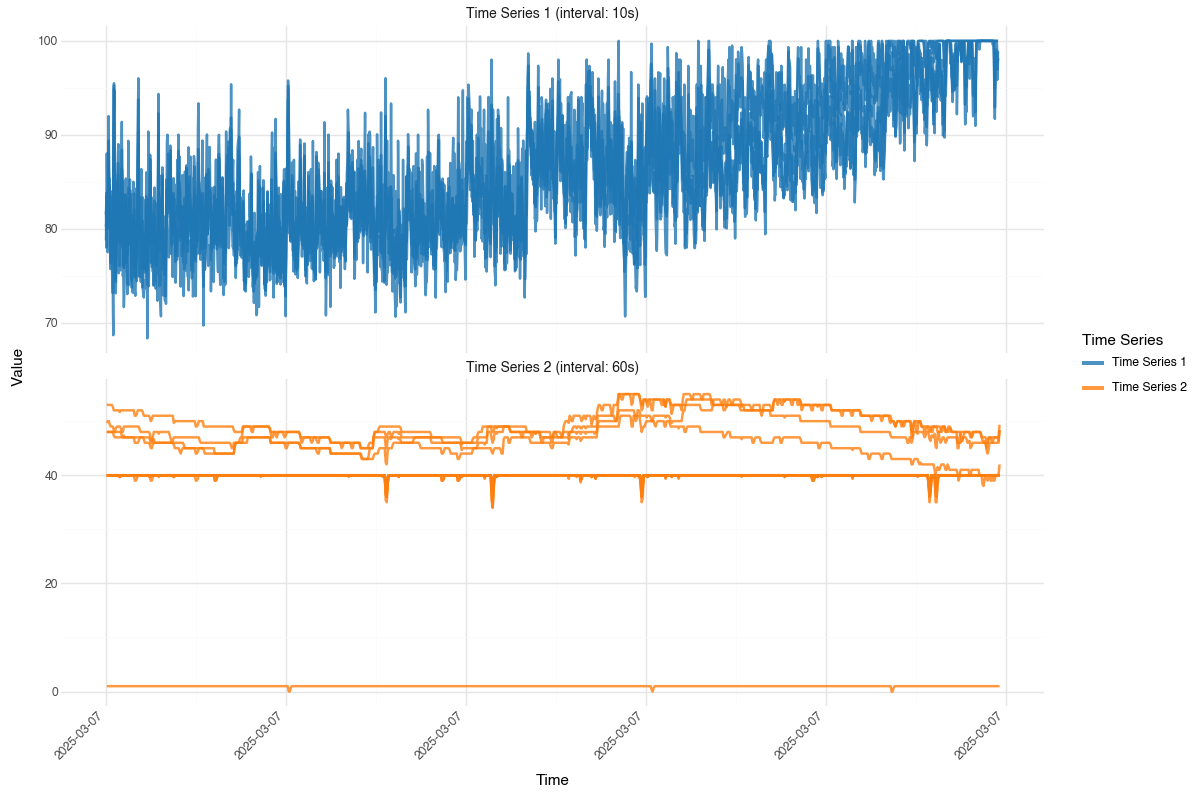}
\end{tcolorbox}

\begin{tcolorbox}[colback=white,colframe=white,boxrule=0pt,arc=0pt,
                  left=0pt,right=0pt,top=0pt,bottom=0pt] 
\begin{tcbraster}[
  raster columns=2,
  raster column skip=8pt,
  raster left skip=0pt,
  raster right skip=0pt,
  raster equal height=rows,
  raster width=\linewidth
]
\begin{answerbox}{red}
\textbf{Model answer:} The anomaly in time-series 1 is a lagging indicator of the anomaly in time-series 2.
\end{answerbox}
\begin{answerbox}{green}
\textbf{Correct answer:} The anomaly in time-series 1 is a leading indicator of the anomaly in time-series 2.
\end{answerbox}
\end{tcbraster}
\end{tcolorbox}

\begin{casebox}[title={Model reasoning}, colback=red!6, colframe=red!40]{Model reasoning}
\textbf{Gemini 3 Pro Preview (VL):} First, ... Time-series 1 (queue utilization) shows a shift in behavior starting around 07:00, with values increasing, but the critical anomaly—hitting 100\% saturation—occurs later, becoming fully established between 11:00 and 12:00. Time-series 2 (available replicas) initially scales up around 07:00 to match the load, but then exhibits an anomalous decrease starting around 09:30, dropping significantly by 11:00 despite the queue not being empty. Comparing the timestamps, the drop in replicas (Time-series 2) begins at ~09:30, which precedes the queue hitting 100\% utilization (Time-series 1) at ~11:30-12:00. Physically, a reduction in consumers (replicas) would cause the queue to fill up (utilization saturation). Since the anomaly in Time-series 2 (the drop) happens before and likely causes the anomaly in Time-series 1 (the saturation), Time-series 1 is the effect and Time-series 2 is the cause. Therefore, the anomaly in Time-series 1 follows the anomaly in Time-series 2.
\end{casebox}

\begin{casebox}[title={Correct reasoning}, colback=green!6, colframe=green!40]{Correct reasoning}
We see average utilization start to trend upwards, triggers an increase in available replicas. However, it is the continued increase in utilization that causes the existing replicas to crash, further reinforcing the saturation of utilization. We note that prior knowledge that scaling Kubernetes deployments react to metrics such as utilization is important here.
\end{casebox}

\end{outercasebox}

\caption{Gemini 3 Pro correctly perceives the graph, but reverses the cause and effect, which is a lack of domain knowledge error.}
\label{fig:ea-4}

\end{figure}

\begin{figure}[hp]
\centering

\begin{outercasebox}{Case Study: Incorrect Perception}

\begin{casebox}[title={Question}, colback=black!2]{Question}
In the following time-series, what is the start time of the anomaly, if an anomaly exists?

Time-series: Top 25 Kafka topic partitions with the highest maximum time lag, filtered by datacenter and specific consumer groups, excluding certain topics and traffic lanes. Results are ordered by mean time lag in descending order.
\end{casebox}

\begin{tcolorbox}[colback=black!2,colframe=black!2,boxrule=0pt,
                  left=0pt,right=0pt,top=0pt,bottom=0pt,
                  height=0.38\textheight,valign=center]
\centering
\includegraphics[height=\tcbtextheight,width=\linewidth,keepaspectratio]{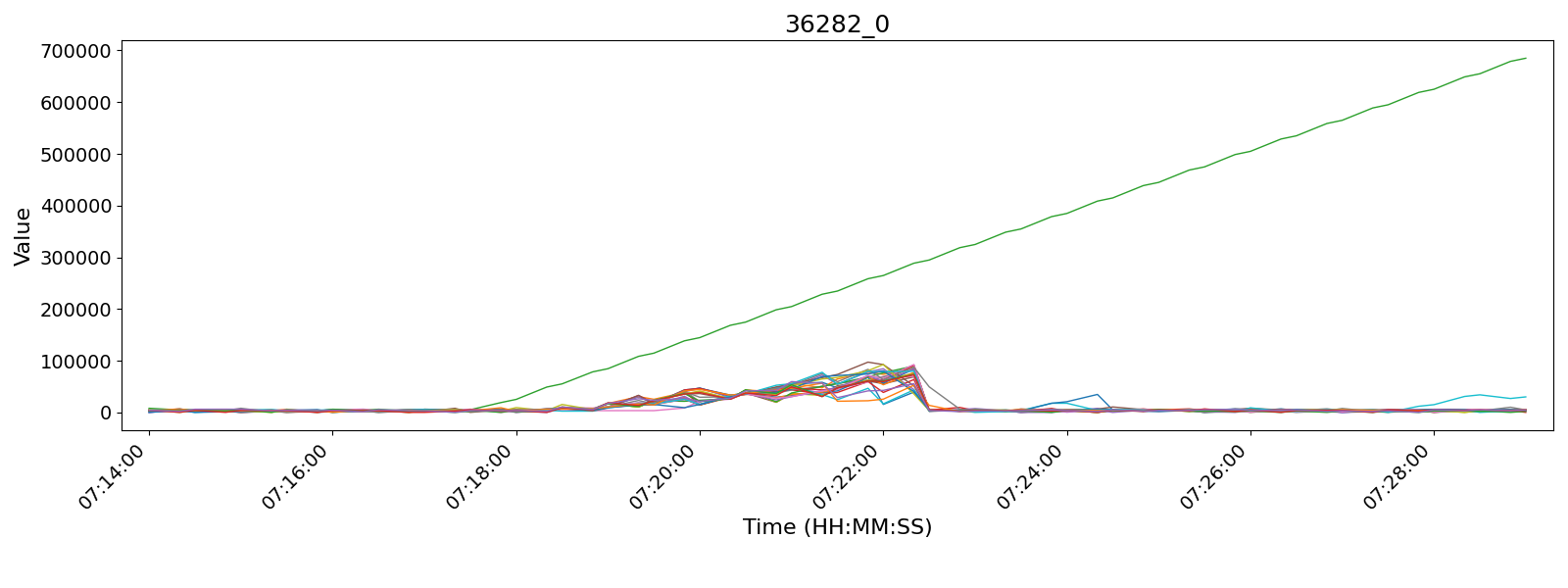}
\end{tcolorbox}

\begin{tcolorbox}[colback=white,colframe=white,boxrule=0pt,arc=0pt,
                  left=0pt,right=0pt,top=0pt,bottom=0pt] %
\begin{tcbraster}[
  raster columns=2,
  raster column skip=8pt,
  raster left skip=0pt,
  raster right skip=0pt,
  raster equal height=rows,
  raster width=\linewidth
]
\begin{answerbox}{red}
\textbf{Model answer:} No Anomaly
\end{answerbox}
\begin{answerbox}{green}
\textbf{Correct answer:} 2025-03-18 07:20:00
\end{answerbox}
\end{tcbraster}
\end{tcolorbox}

\begin{casebox}[title={Model reasoning}, colback=red!6, colframe=red!40]{Model reasoning}
\textbf{Toto-1.0-Qwen3 32B (TSFM-LLM):} - The plot shows 25 Kafka partitions’ maximum time lag, with most series staying near zero and only a few having small, steady values; there is no sudden jump, step change, or coordinated spike across the window. 
\begin{itemize}
    \item The lines that are elevated remain in the same range throughout the period and do not exhibit a new or abrupt deviation at 07:15, 07:20, or 07:25. 
    \item Because the behavior is consistent and stable across time and series, there is no clear anomaly onset within the shown window, so the correct answer is No Anomaly.
\end{itemize}
\end{casebox}

\begin{casebox}[title={Correct reasoning}, colback=green!6, colframe=green!40]{Correct reasoning}
The increasing green channel shows a large increase in maximum time lag of a Kafka topic partition, which indicates an anomaly.
\end{casebox}
\end{outercasebox}

\caption{Toto-1.0-Qwen3 32B does not correctly perceive the time series, missing the changing trend in the green channel.}
\label{fig:ea-5}

\end{figure}

We highlight several case studies (Figures \ref{fig:ea-2}, \ref{fig:ea-3}, \ref{fig:ea-4}, \ref{fig:ea-5}) and present them below.

\subsection{Inter-Annotator Agreement between Users}
\label{aop:iaa}
\begin{table}[t]
\centering
\resizebox{0.7\textwidth}{!}{%
\begin{tabular}{lcc} 
\toprule
Model& Cohen's Kappa (\uparrow)& Krippendorff's Alpha (\uparrow)\\
\midrule
 Domain Expert 1 and Expert 2 &0.667& 0.672\\
 Non-domain Expert 1 and Non-domain Expert 2 & 0.627& 0.626\\
 Domain Expert 1 and Non-domain Expert 1& 0.674&0.673\\
 Domain Expert 1 and Non-domain Expert 2& 0.666&0.667\\
 Domain Expert 2 and Non-domain Expert 1& 0.700&0.705\\
 Domain Expert 2 and Non-domain Expert 2& 0.711&0.715\\
 \bottomrule
\end{tabular}
}
\vspace{1mm}
\caption{Cohen's Kappa and Krippendorff's alpha values for inter-annotator across domain experts and non-domain experts. We find that inter-annotator agreement is generally substantial for all groups for Cohen's kappa, and tentatively acceptable for most pairs for Krippendorff's alpha.}
\label{tab: iaa}
\end{table} We calculate the inter-annotator agreement using Cohen's Kappa \cite{cohen1960coefficient} and Krippendorff's alpha \cite{krippendorff2011computing}, two commonly used robust statistics for inter-annotator agreement, across all pairs of annotators (Table \ref{tab: iaa}). We find that inter-annotator agreement is generally substantial for Cohen's Kappa across pairs ($0.6\le \kappa\le 0.8)$, and tentatively acceptable ($\alpha \ge 0.667$) for Krippendorff's alpha for all pairs except between the non-domain experts.

\subsection{Qualitative Comparison of Model Errors with Domain Experts}
\label{app:expert-comparison}
Analyzing the specific questions that models and experts incorrectly answer, we find that experts tend to make errors on fine-grained perception and some instruction following errors, while models tend to make errors on hallucinations and utilizing domain knowledge.

\begin{figure}[hp]
\centering

\begin{outercasebox}{Domain Expert versus Top Model Error Distribution 1: Model Domain Knowledge Error}

\begin{casebox}[title={Question}, colback=black!2]{Question}
In the following time-series, what is the start time of the anomaly, if an anomaly exists?

Time-series: Sum of estimated custom metric usage for streaming topic statistics, grouped by metric name.
\end{casebox}

\begin{tcolorbox}[colback=black!2,colframe=black!2,boxrule=0pt,
                  left=0pt,right=0pt,top=0pt,bottom=0pt,
                  height=0.38\textheight,valign=center]
\centering
\includegraphics[height=\tcbtextheight,width=\linewidth,keepaspectratio]{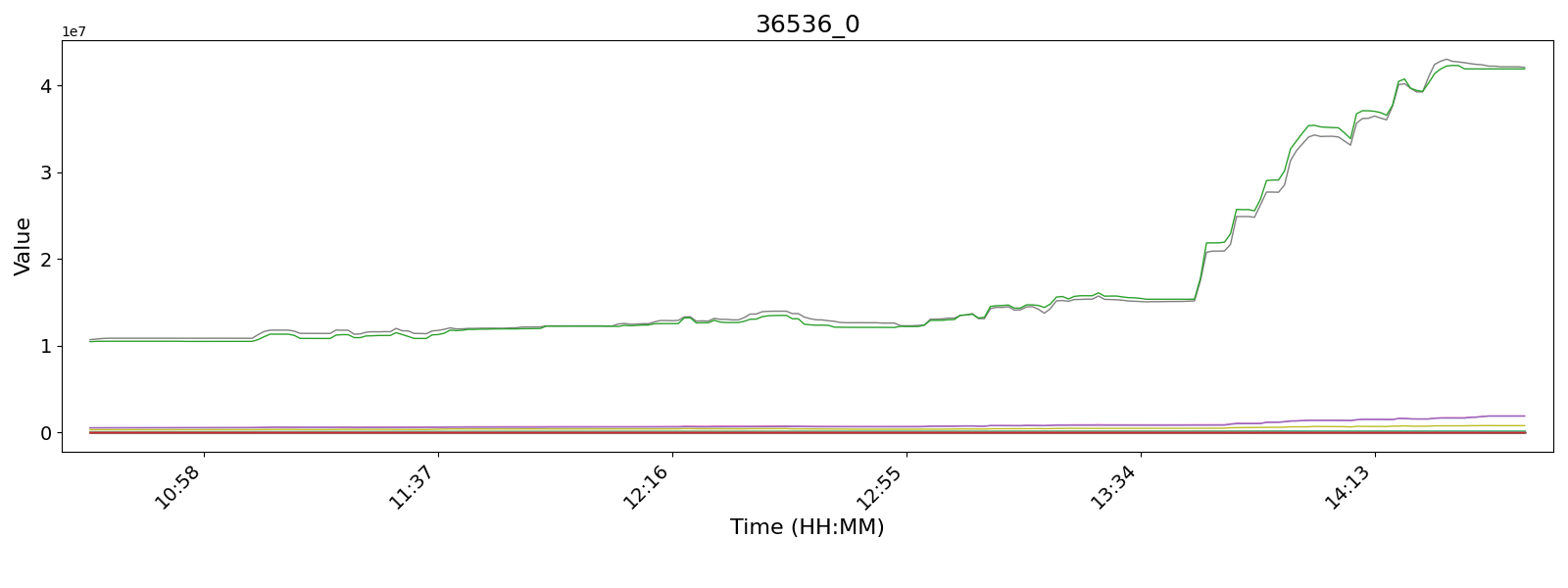}
\end{tcolorbox}

\begin{tcolorbox}[colback=white,colframe=white,boxrule=0pt,arc=0pt,
                  left=0pt,right=0pt,top=0pt,bottom=0pt] %
\begin{tcbraster}[
  raster columns=2,
  raster column skip=8pt,
  raster left skip=0pt,
  raster right skip=0pt,
  raster equal height=rows,
  raster width=\linewidth
]
\begin{answerbox}{red}
\textbf{Model answer:} 2025-03-25 14:00:00
\end{answerbox}
\begin{answerbox}{green}
\textbf{Experts' and Correct answer:} Before the earliest timestamp
\end{answerbox}
\end{tcbraster}
\end{tcolorbox}

\begin{casebox}[title={Model reasoning}, colback=red!6, colframe=red!40]{Model reasoning}
\textbf{GPT-5: }
All channels are fairly stable until around 14:00, after which there is a sudden, sustained upward shift and stepwise increases (especially in the green/grey series), indicating the beginning of the anomaly. Later times (14:30, 14:38) are within the ongoing anomalous period, not its start.
\end{casebox}

\begin{casebox}[title={Correct reasoning}, colback=green!6, colframe=green!40]{Correct reasoning}
Based on domain knowledge, the topic statistics metric should not have such high value in the green and gray channels, as seen relative to other channels. This elevated value is present in the entire segment shown in the time series, which indicates the anomaly started before the timestamps given.
\end{casebox}
\end{outercasebox}

\caption{Top models fail to use sufficient domain knowledge, in contrast to domain experts.}
\label{fig:expert-vs-model-1}

\end{figure}

\begin{figure}[hp]
\centering

\begin{outercasebox}{Domain Expert versus Top Model Error Distribution 2: Expert Fine-grained Perception Error}

\begin{casebox}[title={Question}, colback=black!2]{Question}
In the following time-series, when has the anomaly recovered, if an anomaly exists?

Time-series: Count of successful jobs processed by a notification platform worker, filtered by datacenter and source service, and grouped by job type and organization ID.
\end{casebox}

\begin{tcolorbox}[colback=black!2,colframe=black!2,boxrule=0pt,
                  left=0pt,right=0pt,top=0pt,bottom=0pt,
                  height=0.38\textheight,valign=center]
\centering
\includegraphics[height=\tcbtextheight,width=\linewidth,keepaspectratio]{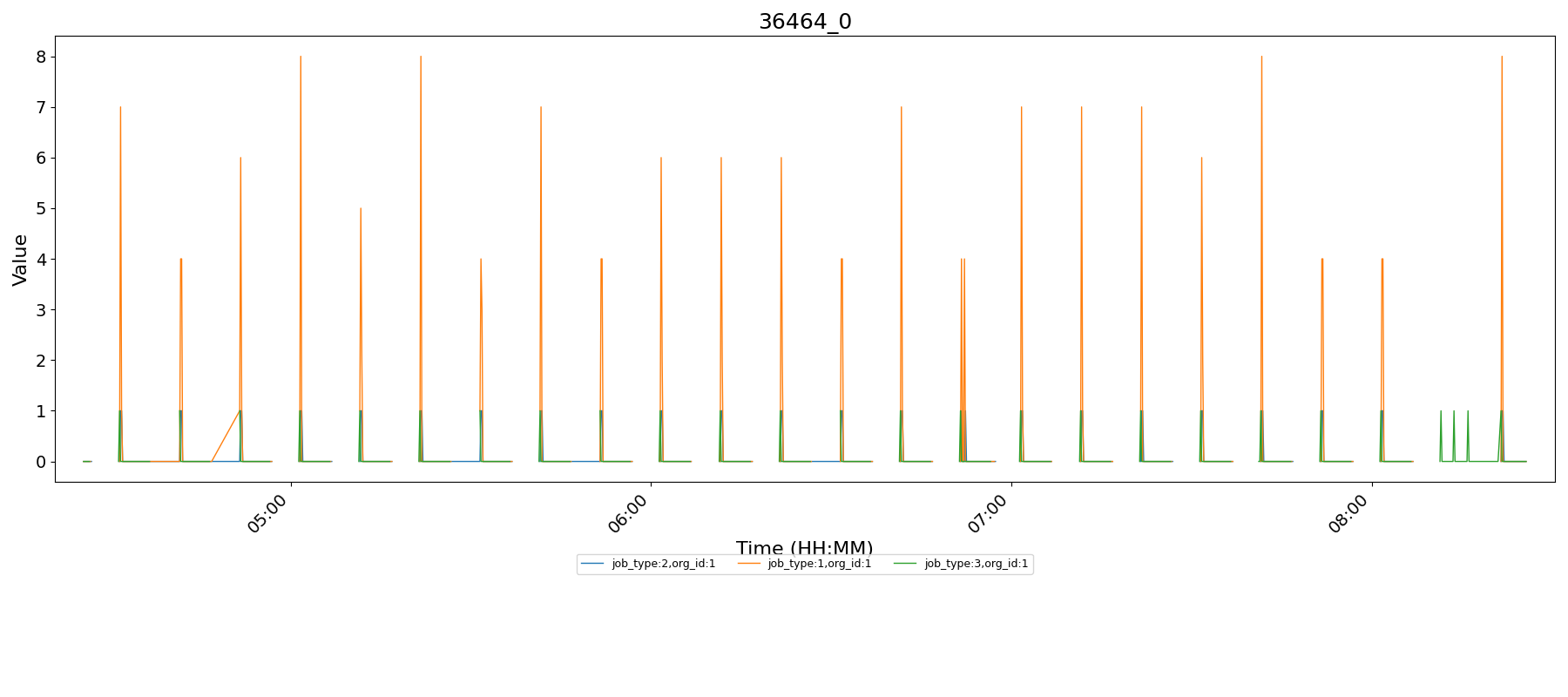}
\end{tcolorbox}

\begin{tcolorbox}[colback=white,colframe=white,boxrule=0pt,arc=0pt,
                  left=0pt,right=0pt,top=0pt,bottom=0pt] %
\begin{tcbraster}[
  raster columns=2,
  raster column skip=8pt,
  raster left skip=0pt,
  raster right skip=0pt,
  raster equal height=rows,
  raster width=\linewidth
]
\begin{answerbox}{red}
\textbf{Experts' answer:} 2025-03-24 08:25:00
\end{answerbox}
\begin{answerbox}{green}
\textbf{GPT-5, Correct answer:} 2025-03-24 08:28:00
\end{answerbox}
\end{tcbraster}
\end{tcolorbox}

\begin{casebox}[title={Model reasoning}, colback=green!6, colframe=green!40]{Model reasoning}
\textbf{GPT-5:} 
An anomaly is present in the green series (job type:3, org id:1): its regular spikes of about 1 drop to near-zero for several minutes around 08:24–08:27, while the other series behave normally. The pattern returns to its normal 1-count spikes at 08:28, indicating recovery at that time.
\end{casebox}

\begin{casebox}[title={Correct reasoning}, colback=green!6, colframe=green!40]{Correct reasoning}
The resolution of the plot makes it difficult for humans and models alike to answer this question, however, models can be more accurate at perceiving these details.
\end{casebox}
\end{outercasebox}

\caption{Experts do not always perceive fine-grained details about exact timings of when anomalies start or end, leading to perceptual errors, in contrast to models.}
\label{fig:expert-vs-model-2}

\end{figure}

\begin{figure}[hp]
\centering

\begin{outercasebox}{Domain Expert versus Top Model Error Distribution 3: Expert Understanding/Instruction-Following Error}

\begin{casebox}[title={Question}, colback=black!2]{Question}
In the following time-series, does the anomaly in this time-series correlate with the anomaly in the other time-series, if anomalies exist?

Time-series 1: Count of build failures for a specific event context provider service in a production datacenter, grouped by context type.

Time-series 2: Rate of indexer errors in an orchestration system for a specific release group, filtered by various dimensions and grouped by pod.
\end{casebox}

\begin{tcolorbox}[colback=black!2,colframe=black!2,boxrule=0pt,
                  left=0pt,right=0pt,top=0pt,bottom=0pt,
                  height=0.38\textheight,valign=center]
\centering
\includegraphics[height=\tcbtextheight,width=\linewidth,keepaspectratio]{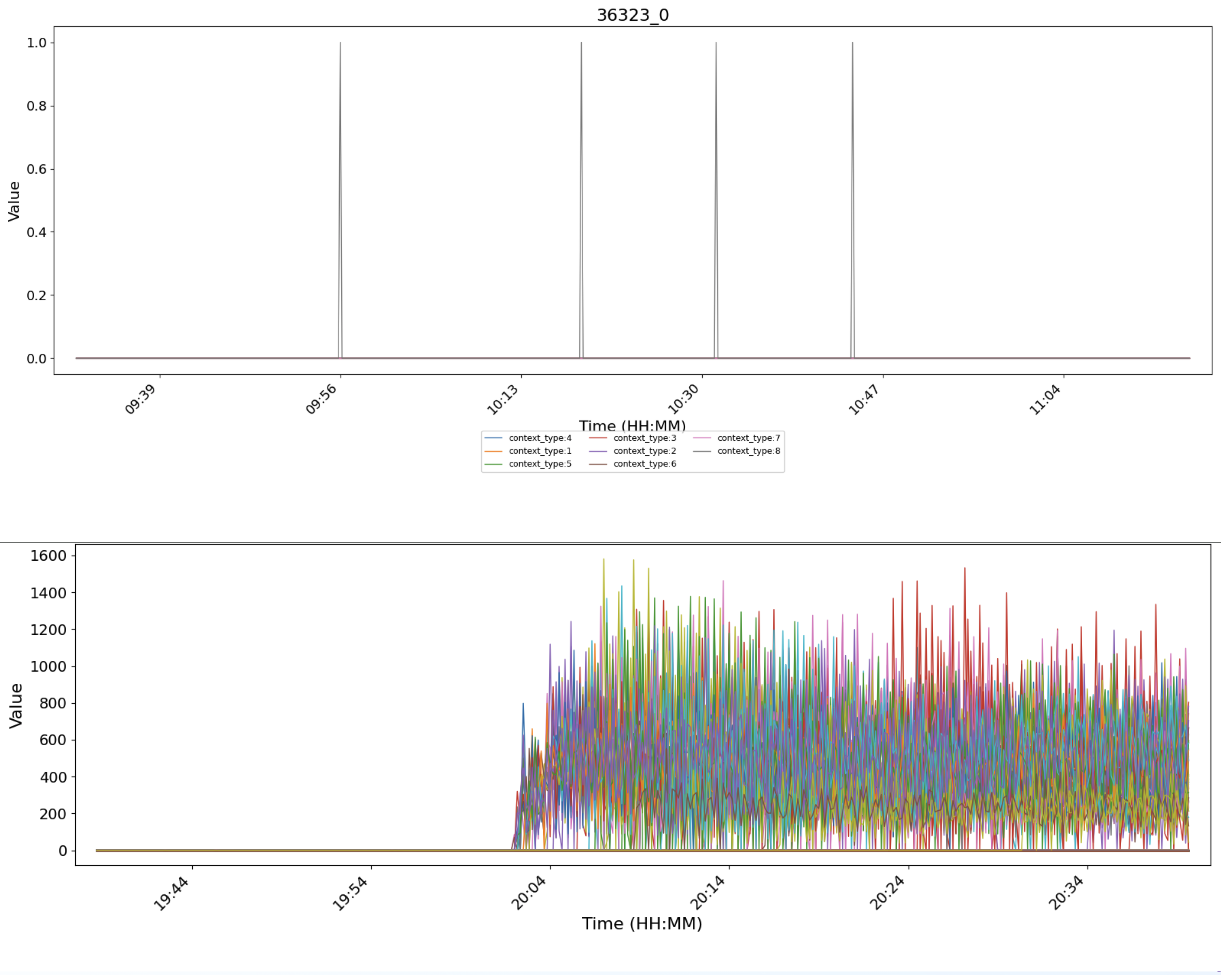}
\end{tcolorbox}

\begin{tcolorbox}[colback=white,colframe=white,boxrule=0pt,arc=0pt,
                  left=0pt,right=0pt,top=0pt,bottom=0pt] %
\begin{tcbraster}[
  raster columns=2,
  raster column skip=8pt,
  raster left skip=0pt,
  raster right skip=0pt,
  raster equal height=rows,
  raster width=\linewidth
]
\begin{answerbox}{red}
\textbf{Expert Answer 1:} No, there is an anomaly only in time-series 2

\textbf{Expert Answer 2:} Yes, there is an anomaly in both and they are correlated
\end{answerbox}
\begin{answerbox}{green}
\textbf{GPT-5, Toto-1.0-QA-Experimental, and Correct answer:} No, there is an anomaly in both but they are not correlated
\end{answerbox}
\end{tcbraster}
\end{tcolorbox}

\begin{casebox}[title={Model reasoning}, colback=green!6, colframe=green!40]{Model reasoning}
\textbf{Toto-1.0-QA-Experimental 32B (TSFM-VLM):} 
Time-series 1 shows a single isolated spike on 2025‑03‑20, while time-series 2 has a large spike on 2025‑03‑25; these occur on different dates and do not overlap. The third plot has a clear surge beginning around 19:59 that grows into a sustained high‑variance burst from \~20:04 to \~20:35, which is a classic ramp‑up followed by a plateau. Because the spikes are separated by days and have different shapes (isolated vs ramping), there is no temporal alignment or co‑movement between the series, so both have anomalies but they are not correlated.
\end{casebox}

\begin{casebox}[title={Correct reasoning}, colback=green!6, colframe=green!40]{Correct reasoning}
The time series displayed are on different days and are temporally unrelated (there is another plot showing no temporal overlap that is elided due to space constraints). However, they both display anomalies relevant to their respective metrics (failures and errors). The TSFM-VLM hallucinates some details but gets the reasoning correct.
\end{casebox}
\end{outercasebox}

\caption{Experts also make some understanding and/or instruction-following errors which models often do not make.}
\label{fig:expert-vs-model-3}

\end{figure}

\begin{figure}[hp]
\centering

\begin{outercasebox}{Domain Expert versus Top Model Error Distribution 4: Model Understanding/Perception Error}

\begin{casebox}[title={Question}, colback=black!2]{Question}
Is the anomaly in time-series 1 a leading or lagging indicator of the anomaly in time-series 2, if anomalies exist?

Time-series 1: Average number of assigned partitions for a specific consumer in a targeted environment, grouped by partition and Kafka consumer topic, with null values filled.

Time-series 2: Sum of Kafka network request rates for specific request types in a targeted cluster and datacenter, grouped by request type and cluster.
\end{casebox}

\begin{tcolorbox}[colback=black!2,colframe=black!2,boxrule=0pt,
                  left=0pt,right=0pt,top=0pt,bottom=0pt,
                  height=0.38\textheight,valign=center]
\centering
\includegraphics[height=\tcbtextheight,width=\linewidth,keepaspectratio]{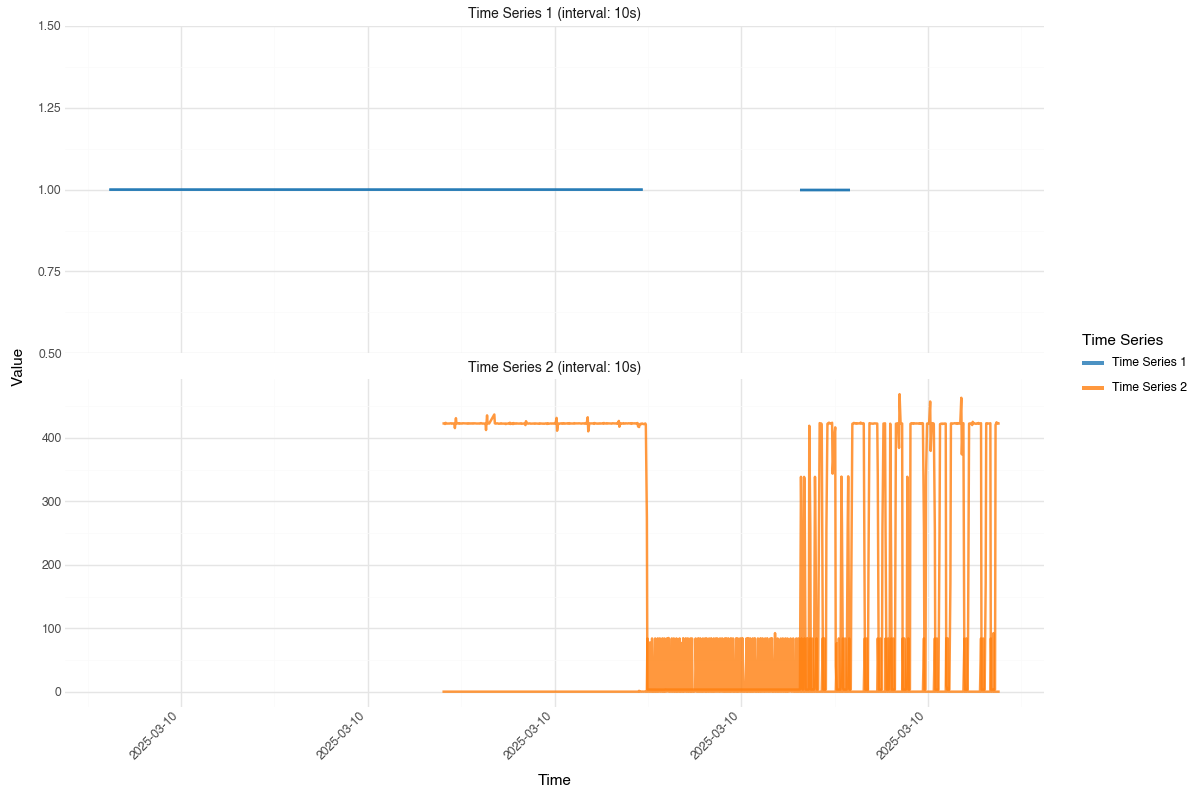}
\end{tcolorbox}

\begin{tcolorbox}[colback=white,colframe=white,boxrule=0pt,arc=0pt,
                  left=0pt,right=0pt,top=0pt,bottom=0pt] %
\begin{tcbraster}[
  raster columns=2,
  raster column skip=8pt,
  raster left skip=0pt,
  raster right skip=0pt,
  raster equal height=rows,
  raster width=\linewidth
]
\begin{answerbox}{red}
\textbf{Model answer:} No anomaly in one or both series
\end{answerbox}
\begin{answerbox}{green}
\textbf{Experts' and Correct answer:} The anomaly in time-series 1 is perfectly correlated to the anomaly in time-series 2.
\end{answerbox}
\end{tcbraster}
\end{tcolorbox}

\begin{casebox}[title={Model reasoning}, colback=red!6, colframe=red!40]{Model reasoning}
\textbf{GPT-5:}
Time-series 1 stays flat at 1.0 throughout the window with no deviations, while time-series 2 shows a large level shift around 14:32 and subsequent spikes. Because time-series 1 exhibits no anomaly, there is no basis to establish a leading or lagging relationship.
\end{casebox}

\begin{casebox}[title={Correct reasoning}, colback=green!6, colframe=green!40]{Correct reasoning}
The model correctly perceives time series 1 to be flat but does not perceive the gaps in the data, which are anomalous drops due to a system failure and a lack of partitions perfectly correlates to the drop in Kafka request rates.
\end{casebox}
\end{outercasebox}

\caption{The model reasons incorrectly, likely due to a lack of contextualization with the time series description.}
\label{fig:expert-vs-model-4}

\end{figure}

We present five example questions that describe some of the differences in errors that are made between models and domain experts in Figures \ref{fig:expert-vs-model-1}, \ref{fig:expert-vs-model-2}, \ref{fig:expert-vs-model-3}, \ref{fig:expert-vs-model-4}.%

\newpage


\newpage

\section{Prompts}
In this section we include the prompts used in benchmark choice generation and evaluation.
\subsection{Time Series Summarization Prompt}
This prompt is used to summarize query strings, or unique database identifiers, to the time series in each incident. We have removed specific sensitive examples or identifiable information in this prompt.
\begin{verbatim}
## Task Description
You are performing a caption generation task for XXXXXX queries.
Your goal is to generate clear, concise, and general captions 
that describe the intent of a given query without leaking sensitive
information about XXXXXXX's architecture, services, environments, or regions.

## Query Format
A standard XXXXXX subquery follows this format:
"operation:metric_name{{filters}}by{{group_by_tags}}"
Queries can consist of a single subquery (simple case) or 
multiple subqueries combined (e.g., ratios, differences, aggregations).


## Handling Service-Specific Metrics
Some queries may contain service-specific metric names referring
to internal XXXXXX services. These should be generalized in the caption.
Below are some examples of such services and how to abstract them:
...
When encountering such service-specific terms in a query,
do NOT mention the service name.
Instead, describe its function in a generic way.
Use `an' instead of `the' when referring to them.

## Caption Generation Guidelines
Summarize the query's purpose in a clear and structured way.
- Describe what is being measured (e.g., memory usage, CPU load, request count)
and how it is processed (e.g., ratio, sum, average).
- DO NOT exposing specific service names, regions, namespaces, env type 
(e.g. dev, staging) or internal identifiers.
Instead, refer to broader categories like "selected workloads",
"target environment", or "specific platform".
- Include filtering and grouping context  
(e.g., "excluding certain workloads" instead of listing exact exclusions).
- Adjust filtering and grouping precision dynamically:
If there are few tags, list them explicitly, otherwise use a more general description
- Maintain clarity and conciseness, ensuring captions are useful
for engineers without unnecessary details.
- Follow the output format strictly.

## Instructions
1. Read the query carefully
2. Follow the Caption Generation Guidelines
3. Generate a caption that accurately summarizes the query while preserving privacy

## Inputs
Query: "{query}"

## Output
The output should be formatted as a JSON instance that
conforms to the JSON schema below.

Here is the output schema:
{
    "query": "<query1> (REQUIRED)",
    "query_id": "<query_id1> (REQUIRED)",
    "caption": "<caption1> (REQUIRED)",
}

## Examples
...

\end{verbatim}
\subsection{Benchmark Options Generation Prompt}
\label{app: option-generation-prompt}
In the prompt provided below, we replace the real incident with an abridged fabricated incident in order to prevent sharing sensitive data. We also include only the first three few-shot examples provided in the prompt for brevity.
\begin{verbatim}
## Task Description
You are performing a question-answering option generation and answer task
for a given set of questions and answers.
Your goal is to generate up to 8 options for each question (including the
correct answer), and label the correct answer.

## Question Format
Each question either has a single query, or two paired queries.

## Question Categories
There are 8 categories of questions:
1. Anomaly Presence

The anomaly presence question is a yes/no question that asks whether an anomaly
is present in the time-series given.
An anomaly is present if the time-series has a value that is significantly
different from the counterfactual values.
Do not generate options for this question, only identify the correct answer.

2. Anomaly Identification

The anomaly identification question asks the user to identify the channel of
the anomaly in the time-series data, if an anomaly exists.
Use the options given in Existing options to generate the answer choices by
including an answer choice for all single options, pairs of options, and triples of options.
DO NOT use other channels other than the existing options. Generate exactly
8 options (7 from the existing options and 1 "No Anomaly" option).
Include a "No Anomaly" option in the answer choices.

3. Anomaly Start

The anomaly start question asks the user to identify the start time of the anomaly
in the time-series data, if an anomaly exists.
Do not include a timestamp that falls outside of the snapshot time range.
If the anomaly is ongoing since the start of the snapshot,
the correct answer should be "Before the earliest timestamp".
The answer choices should be consistent with the time-scale of the snapshot.

Use only the snapshot_png_url and the given time range to generate the answer choices,
not the incident data.
Generate 6 options for this question.
The answer choices should only consist of datetime strings.
Include a "Before the earliest timestamp" option
and a "No Anomaly" option in the answer choices.

4. Anomaly End

The anomaly end question asks the user to identify the end time of the anomaly
in the time-series data, if an anomaly exists. Do not include a timestamp that
falls outside of the snapshot time range. If the anomaly is ongoing at the 
end of the snapshot, the correct answer should be "Not resolved".

Use only the snapshot_png_url and the given time range to generate
the answer choices, not the incident data. Generate 6 options for this question.
The answer choices should only consist of datetime strings.
Include a "Not resolved" option and a "No Anomaly" option in the answer choices.

5. Anomaly Magnitude

The anomaly magnitude question asks the user to identify the
magnitude of the anomaly in the time-series data, if an anomaly exists.
The magnitude is the maximum ratio of the anomaly values to the 
counterfactual non-anomalous values. Here, the magnitudes for the answer choices
should be on a logarithmic scale, in the base that is most natural for the data.
If the counterfactual values are 0, use the absolute deviation
from the mean counterfactual values. Generate 8 options for this question.
Include a "No Anomaly" option in the answer choices.
The answer choices should be unique and distinct from each other.

6. Anomaly Categorization

The anomaly categorization question asks the user to identify the category of the anomaly
in the time-series data, if an anomaly exists.
There are 6 categories:
- Level Shift. This is when the time-series has a major sustained change
in mean value compared to its counterfactual values.
- Transient Spike. This is when the time-series has a sudden spike in value,
but the value returns to the normal range after a very short period of time
with no intervention.
- Change in Seasonality. This is when the time-series has a change
in the seasonal pattern of the data.
- Change in Variance. This is when the time-series has a major sustained change
in the variance of the data.
- Change in Trend. This is when the time-series has a major change
in the trend (long term increase or decrease).
- No Anomaly

Generate 6 options for this question, with the correct answer being one of the categories.

7. Anomaly Correlation

The anomaly correlation question is a paired query question that asks the user
to identify whether the anomalies in two time-series are correlated. 
Use the text from the incident data and the timing of the anomalies 
within the snapshot and incident timeline to identify the correct answer.
There is no need to generate options for this question,
just to identify the correct answer.

8. Anomaly Indicator

The anomaly indicator question is a paired query question that asks the user to
identify whether some anomaly in the first time-series is
a leading or lagging indicator of the anomaly in the second time-series.
Use the text and timestamps from the incident data as well as the
order and timing of the anomalies in the snapshots to identify the correct answer.
There is no need to generate options for this question,
just to identify the correct answer.

## Instructions
1. Read the question and context carefully
2. Use the context to generate up to 8 options for the question.
3. The options should be plausible and realistic,
and should be based on the context
4. The correct option should be one of the options
5. The options should be unique and distinct from each other

## Inputs
Question: <question>
This will be a question from the question categories above.

Formula: <formula>
This will be the internal company formula for the time-series data.
Utilize the keywords in the formula to understand
what the general behavior of the time-series should be.

Snapshot url: <snapshot_url>
This will be the URL to the snapshot image of the time-series data.
Within the snapshot, the x-axis represents the time,
which can vary in time-scale (e.g. minutes, hours, days, etc.). 
The y-axis represents the value of the time-series, which is
labeled with a specific value on the far left side. 

Incident data: <incident_data>
This will be the incident data for the incident that the question is about.
The incident data is a markdown formatted string
that has timestamped messages from the incident, 
which message text that is relevant to the snapshot and the time-series.
The text of the message often informs whether the 
time-series is anomalous, its severity. 
The times at which different time-series are sent
are often relevant to determining  whether they are correlated or lagging or leading.

Existing options: <existing_options>
This will be a list of existing options for the question.

Time range: <from_ts> to <to_ts>
This will be the time range of the snapshot, in a pandas datetime format.
Use this time range to generate the answer choices for anomaly start and end questions.

The response MUST be a JSON in this format.
Respond ONLY with the JSON. All options should be strings.

Output format:
{
    "options": ["<option_A>", "<option_B>", "<option_C>",
    "<option_D>", "<option_E>", "<option_F>", "<option_G>", "<option_H>"],
    "correct_option": "<correct_option>"
}

## Examples 
Incident data for the following questions:
# Incident XXXXX: 
## High-level details
# **Summary:**
# **Root cause: Unknown**
# **Severity: SEV-X**
# **Detected at: 03-01-2025 XX:XX:XX**
# **Customer Impact Description:** 
# **Customer Impact Start:**
# **Customer Impact End:**
## Timeline
# >**[Person A at 2025-03-01]:**
# Some pods are OOMing <time series 1>
# >**[Person B at 2025-03-01]:**
Seeing whether any recent deployments might have to do with this.
# >**[Person C at 2025-03-01]:**
Trying to scale.
...
# >**[Person A at 2025-03-01]:**
Ok, here's the PR.
# > **[Incident state update at 2025-03-01]:** 
Incident status changed from stable to resolved.

Question: In the following time-series,
does the time-series exhibit an anomaly in the given time range?
Time-series: Sum of unavailable replicas for Kubernetes deployments
in a specific datacenter and namespace, focusing on selected applications.
Formula: <Formula>
Snapshot URL: <Snapshot URL>
Existing options: ["Yes", "No"]
Output:
{
    "options": ["Yes", "No"],
    "correct_option": "Yes"
}
Explanation: The incident timeline tells us for the 
given snapshot that pods are hitting resource limits
with out of memory errors, which indicates an anomaly.
Furthermore, within the snapshot, we see that the
unavailable replicas are increasing steadily from 0 to around 30.

Question: In the following time-series, what is the start time of the anomaly,
if an anomaly exists?
Time-series: Sum of unavailable replicas for Kubernetes deployments 
in a specific datacenter and namespace, focusing on selected applications.
Existing options: []
Output:
{
    "options": ["2025-03-01 10:55:00", "2025-03-01 11:05:00", 
    "2025-03-01 11:40:00", "2025-03-01 11:11:40", 
    "Before the earliest timestamp", "No Anomaly"],
    "correct_option": "2025-03-01 11:11:40"
}

Explanation: Here, we use the given time range and crucially the
snapshot image's x-axis to determine the duration of the anomaly,
from when it starts to rise (11:11:40) from 0 to the end of the time range
on the snapshot image (11:40:40). 
The other options either do not cover the entire anomaly or are not the 
minimum duration between the start and end of the anomaly.

Question: In the following time-series, what is the magnitude of the anomaly?
Time-series: Sum of unavailable replicas for Kubernetes
deployments in a specific datacenter and namespace,
focusing on selected applications.
Existing options: []
Output:
{
    "options": ["0.1", "1", "5", "10", "25", "100", "125", "625"],
    "correct_option": "25"
}

Explanation: Here, we use the snapshot image and crucially the y-axis to
determine the magnitude of the anomaly. 
The maximum value is closest to 25 unavailable replicas,
compared to the counterfactual non-anomalous value of 0 
which occurs before the anomaly starts. 
We choose powers of 5 and 10 as the scale of the data because the y-axis
is between 0 and 40, 
and powers of 5 and 10 have many values in this range but 
also outside of this range that can fill the 8 answer choices.
...
\end{verbatim}
\subsection{Prompt for Filtering Benchmark Questions}
\begin{verbatim}
## Task Description
You are filtering questions and answers for a
benchmark dataset based on several key criteria.

## Question Format
Each question either has a single query, or two paired queries.

## Question Categories
There are 8 categories of questions:

1. Anomaly Presence
The anomaly presence question is a yes/no question that
asks whether an anomaly is present in the time-series given.
Key filtering criteria:
- The entire time-series should be visible in the snapshot,
otherwise do not filter out any questions for this category.

2. Anomaly Identification

The anomaly identification question asks the user to identify
the channel of the anomaly in the time-series data, if an anomaly exists.
Key filtering criteria:
- The entire time-series should be visible in the snapshot,
otherwise do not filter out any questions for this category.

3. Anomaly Start

The anomaly start question asks the user to identify the 
start time of the anomaly in the time-series data, if an anomaly exists.
Key filtering criteria:
- The entire time-series should be visible in the snapshot
- The answer choices should not include any time outside 
of the snapshot time range
- If all answer choices are outside of the snapshot
time range, then the question should be filtered out
- The answer choices should be consistent
with the time-scale of the snapshot.

4. Anomaly End

The anomaly end question asks the user to identify the end time
of the anomaly in the time-series data, if an anomaly exists.
Key filtering criteria:
- The entire time-series should be visible in the snapshot
- The answer choices should not include any time 
outside of the snapshot time range
- If all answer choices are outside of the snapshot time range,
then the question should be filtered out
- The answer choices should be consistent 
with the time-scale of the snapshot.

5. Anomaly Magnitude

The anomaly magnitude question asks the user to identify
the magnitude of the anomaly in the time-series data,
if an anomaly exists. The magnitude is the maximum ratio
of the anomaly values to the counterfactual non-anomalous values,
or the maximum absolute deviation from the counterfactual value
if the counterfactual value is 0.
Key filtering criteria:
- The answer choices should include magnitudes that are 
reasonable for the data. If the scale of the data is on the order of 10,
but the options do not include 10, then the question should be filtered out.

6. Anomaly Categorization

The anomaly categorization question asks the user to identify
the category of the anomaly in the time-series data, if an anomaly exists.
There are 6 categories:
- Level Shift
- Transient Spike
- Change in Seasonality
- Change in Trend
- Change in Variance
- No Anomaly

Key filtering criteria:
- Do not filter out any questions for this category.

7. Anomaly Correlation

The anomaly correlation question is a paired query question
that asks the user to identify whether the anomalies in two time-series
are correlated.
Key filtering criteria:
- If the two time-series have completely non-overlapping time ranges,
then the question should be filtered out. 
This can be deduced from the snapshot x-axis labels,
as well as the time stamps in the incident timeline.
However, if the two time-series have overlapping time ranges,
then the question should not be filtered out.

8. Anomaly Indicator

The anomaly indicator question is a paired query question 
that asks the user to identify whether some anomaly in the first time-series is
a leading or lagging indicator of the anomaly in the second time-series.
Key filtering criteria:
- If the two time-series have completely non-overlapping time ranges,
then the question should be filtered out. This can be deduced from the snapshot
x-axis labels, as well as the time stamps in the incident timeline. 
However, if the two time-series have overlapping time ranges,
then the question should not be filtered out.

Key filtering criteria for ALL questions:
- If the snapshot does not give enough information to identify
the time-series data of the specified query, then the question should be filtered out.
- If it is difficult to identify most of the x-axis labels
or time-series points in the snapshot,
then the question should be filtered out. This does not apply if most of the
x-axis is visible but there is one point label that marks a specific time.
- If the answer options are empty, then the question should be filtered out.
- Do not filter out questions for any other reason besides the ones listed above.

## Instructions
1. Read the question and context carefully
2. Use the key filtering criteria to determine if the question should be filtered out.

## Inputs
Question: <question>
Formula: <formula>
Snapshot url: <snapshot_url>
Incident data: <incident_data>

The response MUST be a JSON in this format. Respond ONLY with the JSON.
Do not include any extraneous formatting
(such as newline characters or extra backslashes) or Markdown quotes.
Make sure the response is an integer (either 0 or 1),
1 if the question should be filtered out, 0 if it should not be filtered out.

Output format:
{
    "filtered_out": <filtered_out>
}

## Examples
Incident data for the following questions:
# Incident XXXXX: 
## High-level details
# **Summary:**
# **Root cause: Unknown**
# **Severity: SEV-X**
# **Detected at: 03-01-2025 XX:XX:XX**
# **Customer Impact Description:** 
# **Customer Impact Start:**
# **Customer Impact End:**
## Timeline
# >**[Person A at 2025-03-01]:**
# Some pods are OOMing <time series 1>
# >**[Person B at 2025-03-01]:**
Seeing whether any recent deployments might have to do with this.
# >**[Person C at 2025-03-01]:**
Trying to scale.
...
# >**[Person A at 2025-03-01]:**
Ok, here's the PR.
# > **[Incident state update at 2025-03-01]:** 
Incident status changed from stable to resolved.
Question: In the following time-series, 
what is the start time of the anomaly, if an anomaly exists?
Time-series: Negative sum of application start events for
a specific node group in a production environment,
excluding certain conditions, grouped by host and datacenter.
Snapshot URL: <Snapshot URL>
Options: "['2025-03-05 03:01:00'
 '2025-03-05 03:05:59'
 '2025-03-05 03:07:34'
 '2025-03-05 03:11:44'
 '2025-03-05 03:16:20'
 'Before the earliest timestamp',
 'No Anomaly']"

Output:
{
    "filtered_out": 1
}
Explanation: The start times of all options are outside of the snapshot time range,
so the question should be filtered out.

Question: "In the following time-series, 
does the anomaly in this time-series correlate with the anomaly in the other time-series?
Time-series 1: Rate of gRPC errors in a client metrics catalog
for a specific organization, excluding canceled errors, 
filtered by datacenter and canary status,
and grouped by service, error code, cause, and query source.
Time-series 2: Rate of gRPC requests to a historical data query
service without organization ID, filtered by datacenter, environment,
and specific pod, grouped by error and code."
Option: "['No, there is an anomaly in both but they are not correlated'
 'Yes, there is an anomaly in both and they are correlated'
 'No, there is no anomaly in either time-series'
 'No, there is an anomaly only in time-series 1'
 'No, there is an anomaly only in time-series 2']"

Output:
{
    "filtered_out": 1
}

Explanation: The two time-series have completely non-overlapping time ranges,
so the question should be filtered out.

Question: "In the following time-series, does the anomaly in this
time-series correlate with the anomaly in the other time-series?
Time-series 1: Ratio of average CPU usage to CPU limits for a 
specific processing service in a target environment, 
grouped by Kubernetes deployment and cluster.
Time-series 2: Maximum value of a watermark pod autoscaler controller
for a specific processing service, 
filtered by datacenter and resource namespace,
grouped by cluster and metric name, divided by 1000."
Option: "['No, there is an anomaly in both but they are not correlated'
 'Yes, there is an anomaly in both and they are correlated'
 'No, there is no anomaly in either time-series'
 'No, there is an anomaly only in time-series 1'
 'No, there is an anomaly only in time-series 2']"

Output:
{
   "filtered_out": 0
}

Explanation: The two time-series have the same time range
as we can see from the snapshot, thus the question should not be filtered out.

Question: "In the following time-series, 
which channels are exhibiting anomalies in the given time range, if any?
Time-series: Percentage of query errors in a server query resolution system,
calculated as the ratio of error queries to total queries,
filtered by datacenter and grouped by method, version, gRPC method, and shard."
Option: ""['method:1' 'version:1' 'grpc_method:1' 'shard:1' 'method:1,version:1'
 'method:1,grpc_method:1' 'method:1,shard:1' 'No Anomaly']""

Output:
{
    "filtered_out": 0
}

Explanation: The question should not be filtered out because
the options are valid and the time-series is visible in the snapshot.
\end{verbatim}
\subsection{Few-Shot Prompt for Model Evaluation}
\label{app: test-prompt}
        \begin{verbatim}
## Task Description
You are an expert at analyzing time-series data 
and answering questions about anomalies.
Your task is to answer the given question
about time-series anomalies by selecting the most appropriate option.
Focus on the key aspects of the anomaly being
analyzed and provide a clear explanation for your choice.

Input:
- Question: <question>
- Options: <options>
- Snapshot PNG: URL to a PNG image of the time-series.

## Question Categories
There are 8 categories of questions:
1. Anomaly Presence

The anomaly presence question is a yes/no question
that asks whether an anomaly is present in the time-series given.
An anomaly is present if the time-series has a value that
is significantly different from the counterfactual values.

2. Anomaly Identification

The anomaly identification question asks the user to identify
the channel of the anomaly in the time-series data, if an anomaly exists.
You must identify the correct channels referenced in the options,
and decide based on the meaning of the time series description as well
as the context of the other channels to decide which channel(s) is anomalous.

3. Anomaly Start

The anomaly start question asks the user to identify the start
time of the anomaly in the time-series data, if an anomaly exists.
The start time is the first time the anomaly appears in the time-series.
If there is no exact timestamp for the start time,
the correct answer is the timestamp closest to the start of the anomaly.

4. Anomaly End

The anomaly end question asks the user to identify the end time
of the anomaly in the time-series data, if an anomaly exists.
The end time is the last time the anomaly appears in the time-series.
If there is no exact timestamp for the end time,
the correct answer is the timestamp closest to the end of the anomaly.

5. Anomaly Magnitude

The anomaly magnitude question asks the user to identify the magnitude
of the anomaly in the time-series data, if an anomaly exists.
The magnitude is the maximum ratio of the anomaly values to the
counterfactual non-anomalous values.
Here, the magnitudes for the answer choices are on a logarithmic scale,
in the base that is most natural for the data.
If the counterfactual values are 0, use the absolute
deviation from the mean counterfactual values.

6. Anomaly Categorization

The anomaly categorization question asks the user to identify
the category of the anomaly in the time-series data, if an anomaly exists.
There are 6 categories:
- Level Shift. This is when the time-series has a sustained change 
in mean value.
- Transient Spike. This is when the time-series has a sudden spike
in value, but the value returns to the normal range after a very
short period of time with no intervention.
- Change in Seasonality. This is when the time-series has a change
in the seasonal pattern of the data.
- Change in Variance. This is when the time-series has a major 
sustained change in the variance of the data.
- Change in Trend. This is when the time-series has a major change
in the trend (long term increase or decrease).
- No Anomaly

Generate 4 options for this question, 
with the correct answer being one of the categories.

7. Anomaly Correlation

The anomaly correlation question is a paired query question that
asks the user to identify whether the anomalies in two time-series
are correlated. Two anomalies are correlated if they have a known
causal relation, if the time series have similar trends over time, or if they
have the same underlying root causes.

8. Anomaly Indicator

The anomaly indicator question is a paired query question that asks
the user to identify whether some anomaly in the first time-series is
a leading or lagging indicator of the anomaly in the second time-series.
Use the timing of the anomalies in the images to identify the correct answer.
    
## Answer Format
The answer should match one of the options exactly. 
Do not include the letter of the option in the answer.
Include a detailed explanation of your reasoning for the answer.

The response MUST be a JSON in this format. Respond ONLY with the JSON.
Do not include any extraneous formatting or Markdown quotes.
Output format:
{
    "answer": <answer>,
    "reasoning": <reasoning>
}
\end{verbatim} 
\end{document}